\documentclass[preprint,12pt]{elsarticle}

\usepackage{natbib}
\usepackage{amssymb}
\usepackage{color}
\usepackage{tabularx}
\usepackage{graphicx}
\usepackage{amsmath}
\usepackage{mathtools}
\usepackage{url}
\usepackage{float}
\usepackage[noend,ruled,linesnumbered]{algorithm2e}
\DontPrintSemicolon
\usepackage{caption}

\newcommand{\cA}{{\mathcal A}}
\newcommand{\cX}{{\mathcal X}}
\newcommand{\cD}{{\mathcal D}}
\newcommand{\cR}{{\mathcal R}}
\newcommand{\hbs}{\textsc{HBS}\xspace}
\newcommand{\rsd}{\textsc{RSD}\xspace}
\newcommand{\dsa}{\textsc{DSA}\xspace}
\newcommand{\acls}{\textsc{ACLS}\xspace}
\newcommand{\dsarc}{\textsc{DSA\_RC}\xspace}
\newcommand{\greedy}{\textsc{Greedy}\xspace}
\newcommand{\random}{\textsc{Random}\xspace}

\newtheorem{definition}{Definition}{\bfseries}{\itshape}

\journal{Expert Systems with Applications}
\biboptions{authoryear}

\begin{document}

\begin{frontmatter}

\title{Distributed course allocation with asymmetric friendships}


\author[AU2,CY]{Ilya Khakhiashvili}
\ead{ilk17895@gmail.com}
\author[AU,CY]{Lihi Dery\corref{CORR}}
\ead{lihid@ariel.ac.il} 
\author[AU,CY]{Tal Grinshpoun}
\ead{talgr@ariel.ac.il} 

\cortext[CORR]{Corresponding author}
\address[AU2]{Department of Computer Science, Ariel University, Ariel, Israel}
\address[AU]{Department of Industrial Engineering and Management, Ariel University, Ariel, Israel}
\address[CY]{Ariel Cyber Innovation Center, Ariel, Israel}

\begin{abstract}
Students' decisions on whether to take a class are strongly affected by whether their friends plan to take the class with them. A student may prefer to be assigned to a course they likes less, just to be with their friends, rather than taking a more preferred class alone. It has been shown that taking classes with friends positively affects academic performance. Thus, academic institutes should prioritize friendship relations when assigning course seats. The introduction of friendship relations results in several non-trivial changes to current course allocation methods. This paper explores how course allocation mechanisms can account for \emph{friendships} between students and provide a unique, distributed solution. In particular, we model the problem as an asymmetric distributed constraint optimization problem and develop a new dedicated algorithm. Our extensive evaluation includes both simulated data and data derived from a user study on 177 students' preferences over courses and friends. The results show that our algorithm obtains high utility for the students while keeping the solution fair and observing courses' seat capacity limitations. 
\end{abstract}


\end{frontmatter}

\section{Introduction}
Assigning students to courses is a challenging problem. To do so, we need to consider student preferences -- i.e., which classes they like and which classes they don't -- as well as university requirements such as seat limits. Course allocation mechanisms must account for these constraints, as well as ensure that the resulting allocation satisfies fairness, equality, and welfare goals.

Course allocation deals with indivisible items (courses). 
Allocation of indivisible items is a well-studied problem with applications in various domains. In this problem, a set of \emph{agents} needs to be assigned a set of indivisible \emph{items}. The agents have \emph{subjective utilities} over the items they receive.
We focus on dependencies between agents, i.e., agents' preferences over items also depend on the assignments of other agents.
Relevant settings with agent dependencies include potential neighbors in land allocation and shift allocation problems (of nurses, waiters, etc.), where workers are assigned shifts with their preferred colleagues.
This property has so far been overlooked in research and practice.
Course allocation is a befitting example of this setting:
the agents are university students, and the items are seats in classes. One student might choose to take Introduction to Algorithms, whereas another student might select Machine Learning.  However, if students prefer to take classes with their friends, the intrinsic desirability of a class seat depends on how other class seats are assigned.

For students, being allocated to courses with their friends may be a top priority. 
Students might prefer to be assigned to a less-preferred course with their friends rather than taking a more preferred class alone.
Taking classes with friends has a positive effect on academic performance~\citep{lancieri2017asymmetry, sparrowe2001social, smith2007psst, witkow2010school}. Our definition of friendship is asymmetric: a student is friends with another student if they wish to be assigned to courses with them, regardless of the wishes of the other student.  For example, Alice wishes to study with Bob, so Bob is Alice's friend, but the opposite does not necessarily hold. 
Contrary to other parameters in this problem, information regarding friendships may be sensitive, especially as preferences may be asymmetric, and students' social status can be inferred from them. Thus, students may wish to refrain from reporting their friendship to a central authority (such as the school administrative office).


In this work, we model the allocation of courses to students while considering not only the students' preferences over courses but also their preferences over classmates.
We were inspired by a recent paper that considered land allocation among friends~\citep{elkindPTZ20}. However, that paper deals with single-unit allocation (one piece of land to each agent) and suggests a bidding mechanism. We add friendships to a multi-unit setting and formulate the problem as an asymmetric distributed constraint optimization problem (ADCOP)~\citep{grinshpoun2013asymmetric}. 
We also develop a novel DCOP algorithm that facilitates the specifics of this course allocation problem. 

\medskip
\noindent\textbf{Contributions:}
Our contribution is five-fold:
\begin{itemize}
    \item Multi-unit allocation with friendships -- the proposed allocation explicitly considers friendships.
    \item Decentralization -- our suggested solution is decentralized, which means that agent preferences are distributed and are unknown upfront to a single machine or human coordinator.
    \item Welfare -- our proposed algorithm commonly finds solutions of high utility (social welfare).
    \item Fairness in order -- our algorithm does not apply an ordering of the agents, and, therefore, no explicit or implicit prioritization is given to any of the agents.
    \item Equality -- the solutions provided by our algorithm maintain equality among the agents in terms of their personal utilities.
\end{itemize}

Our suggested algorithm is evaluated against five baseline algorithms. Agent preferences over courses are taken from a real-world data set and from data we collected in an experiment conducted at Ariel University. The evaluation reveals that our algorithm, \dsarc, maintains fairness, provides high utility for the students, and finds valid solutions even in very constrained scenarios. In addition, we discovered that there is a correlation between the strength of the friendship and the performance of our algorithm. This provides a significant advantage to students who prioritize course choices based on friendship over the contents of the course itself.

Throughout the work, we use the terms students and agents interchangeably. 
While we demonstrate our solution for course allocation with friendships, our model is general and can be adapted to solve various multi-allocation problems where it is required to consider friendships.

A preliminary version of this work was presented as a short conference paper~\citep{khakhiashvili2021course}. 
That version considered only simulated data with fewer algorithms and evaluation metrics. Conversely, the evaluation in the present version is much more elaborate and also contains a user study. The present article also provides an expanded view of related works, offering an original perspective on the gap in the existing literature. 

The rest of this work is organized as follows. In Section~\ref{sec:preliminaries}, we overview existing course allocation models and provide background and a formal definition of the ADCOP model. Then, in Section~\ref{sec:solution}, we present our proposed solution, including the modeling of the course allocation problem with friendships, the formulation of the problem as an ADCOP, and the new algorithm. We present an extensive evaluation in Section~\ref{sec:eval} and conclude with a discussion in Section~\ref{sec:conclusion}.

\section{Preliminaries}  
\label{sec:preliminaries}
We begin with relevant related works on course allocation mechanisms (Section \ref{subsec:courses}), including related models, and continue with the background and formal definition of the ADCOP model (Section \ref{subsec:adcop}).

\subsection{Course allocation mechanisms}\label{subsec:courses}
Course allocation is a (multi-unit) one-sided allocation problem, as students have preferences over courses, but courses do not have preferences over students. In some cases~\citep{diebold2014course, nogareda2016optimizing, bichler2021assign}, course instructors may have preferences over the students, such as students' seniority, honor status, prior experience with the subject, thus resembling \textit{two-sided allocations} such as school choice~\citep{abdulkadirouglu2003school}, college admission~\citep{biro2008student}, hospital-residents matching~\citep{roth2002economist}, or refugee resettlement~\citep{delacretaz2016refugee}. 

In this research, we focus on \textit{one-sided allocation}.
Other one-sided assignment problems include assigning workers to shifts, airport landing or takeoff slots to aircraft, task allocation~\citep{wang2022task, zhao2023ppo} and blocks of time or communication bandwidth for shared resources such as NASA's deep space network antennas~\citep{johnston2017ai,hackett2018spacecraft}. 
Course allocation should aim to address the following requirements: 
\begin{itemize}
    \item \textbf{Social welfare optimal}: students will be as satisfied as possible with their allocation (i.e., the mechanism finds a solution with maximum utility). 
    \item \textbf{Pareto optimal}: no student will wish to change their allocation with that of another student. No alternative allocation provides more utility to one or more students without impairing the utility of any student.
    \item \textbf{Fair in order}: the order in which allocations are made is fair. The queue is either in random order or according to an agreed metric such as seniority.
    \item \textbf{Equal}: the dispersion of satisfaction among the students is low. 
    \item \textbf{Incentive compatible}: it is the dominant strategy of all students to report their truthful preferences, regardless of the reports of the other students. This reduces the cognitive load on the students since they need not attempt to predict the preferences of other students. 
\end{itemize}

While mechanisms that comply with the above requirements exist for single-unit allocation under certain conditions (see works that stem from \citet{gale1962college}),
for multi-unit problems, the only solutions which are Pareto efficient and incentive compatible are dictatorships \citep{klaus2002strategy, ehlers2003coalitional, papai2001strategyproof}, meaning that some agent selects courses before the others, thus making the solution extremely unfair. Even for just two agents, truthfulness and fairness are incompatible~\citep{amanatidis2017truthful}. 
Envy-free (EF) allocations are another often studied requirement. 
An EF allocation means that every student prefers their allocation to that of every other student~\citep{foley1966resource}.
Since EF allocations may not exist, weaker fairness notations (or additive relaxations of envy-freeness) are often examined.  One option is envy-freeness up to one good (EF1)~\citep{budish2012multi,Lipton2004EF1}, meaning that an allocation is EF1 if for any two agents $i,j$, there is some item in one $j$'s bundle whose removal results in $i$ not envying $j$. Envy-freeness up to any good (EFX)~\citep{gourves2014near, caragiannis2019unreasonable} and maximin share fairness (MMS)~\citep{budish2011combinatorial} are also studied (see e.g. ~\citep{amanatidis2021allocating}).  

The most simple and widely used solution to the course allocation problem is to allocate courses on a first-come-first-serve basis. However, this solution does not address any of the requirements listed above. 
Mechanisms that address social welfare are the bidding point mechanisms (see e.g., \citep{krishna2008research, kominers2010course, juthamanee2021token}). In these, students are given a bid endowment of points and each student submits a bid vector that allocates these points to courses. Following the submission of all students, the bids are sorted into a single list in descending order and accepted if they result in a feasible allocation. It was shown by~\citet{sonmez2010course} that bids might significantly differ from the students' true preferences, thus, these mechanisms are not incentive compatible. 
Moreover, they do not address the other requirements either. 

A mechanism that addresses social welfare and Pareto optimality was developed by \citet{sonmez2010course}. Each student is required to submit an ordered list of preferences over the courses, as well as bids. Students are allocated according to their highest bids. Students rejected from a course receive their next most preferred course. However, as with the bidding mechanism, this is not incentive compatible, fair, or equal. 

Mechanisms that address fairness in order are the draft mechanisms. In these, students are drawn from a ``draft order'' and submit their preferences over individual courses, one course in each turn. 
\citet{budish2012multi} describe $HBS$, a draft mechanism employed by Harvard Business school. In an iterative process, students are allocated to courses one by one over a series of $m$ rounds. In odd rounds, students are allocated according to a predefined ascending order and in even rounds, students are allocated according to the same order, but this time in descending order. Students 
are allocated to their most preferred courses if this course still has capacity. 
Once the $m$ rounds are completed, an additional multi-pass algorithm is employed, in which students have an opportunity to drop courses they were assigned to and enroll in courses that have capacity. This is performed until no more add-drop requests can be fulfilled. 
As students only submit their ordered preferences, it may be that one student is almost indifferent between two courses, while for the other student, one of these courses will greatly increase their utility. The mechanism does not know of this and thus cannot increase social welfare. 

Incentive compatibility is addressed by random serial dictatorship mechanisms, in which students are sorted randomly and each, in turn, picks an entire bundle of courses~\citep{papai2001strategyproof, ehlers2003coalitional,hatfield2009strategy}. However, fairness is neglected here. 
A different attempt to address incentive compatibility, specifically strategyproofness in the large~\citep{azevedo2019strategy} was made by~\citet{budish2011combinatorial}, with a mechanism that approximates competitive equilibrium from equal incomes ($A-CEEI$). Students are assigned budgets that are approximately (but not exactly) equal. The mechanism is EF1 and has other desirable criteria, but it cannot be implemented in practice as capacity course constraints are not met, and there is no polynomial time algorithm. A practical implementation (titled $CM$) was later suggested by~\citet{budish2017course}, but to use this algorithm correctly, students are required to report their preferences correctly, which is a very hard task~\citep{budish2022can}. 

Measures that address social welfare, fairness, and equity are proposed by~\citet{yekta2020}. They use linear programming to deal with ordinal and then cardinal optimization using two-stage optimization. They find the maximum possible total rank and then maximize the total bid points among all rank-optimal solutions. They perform this two-part optimization once for the whole market or at the end of each round. 

Allocations that are friendship considerate and private are almost  nonexistent. The only work available considered one friend for each agent when allocating land~\citep{elkindPTZ20}. Land allocation is a single-unit allocation (one piece of land to each agent, as opposed to multiple courses to each agent. In that work, the authors identified incentive-compatible settings, i.e., settings that incentivize agents to report their plot values and friendship information truthfully. 

To the best of our knowledge, ours is the first work that addresses friendships in a multi-unit setting. We deal with social welfare, fairness in order, equality, and the validity of solutions. Other properties, such as incentive compatibility and Pareto optimality, are out of the scope of this study and may be handled in future work. 

\subsection{Asymmetric DCOP} \label{subsec:adcop}
DCOP~\citep{modi2005adopt,fioretto2018distributed} constitutes a powerful framework for describing distributed optimization problems in terms of constraints, where the constraints are enforced by distinct participants (agents). In the past two decades, various algorithms for solving DCOPs were developed and then applied to solving problems in practical domains, such as meeting scheduling~\citep{maheswaran2004taking,gershman2008scheduling}, mobile sensor nets~\citep{zhang2005distributed,farinelli2008decentralised}, vehicle routing~\citep{leaute2011distributed}, and the Internet of Things~\citep{lezama2019agent}. However, the standard DCOP model fails to capture scenarios in which the interactions (constraints) between the agents are asymmetric. In this study, we consider friendships between students (agents), which is, in many cases, a non-symmetric relation (cf.~\citep{carley1996cognitive}). Consequently, we adopt the ADCOP model, which enables asymmetric relations between the agents~\citep{grinshpoun2013asymmetric}.


Formally, an ADCOP is a tuple $\langle\cA,\cX,\cD,\cR\rangle$ where $\cA = \{1, 2 , \ldots, n\}$ is a finite set of agents; $\cX = \{X_1, X_2, \ldots , X_n\}$ is a finite set of variables, each held by a single agent;
\footnote{The DCOP model enables an agent to hold more than one variable, but this feature is not needed in our context. Therefore, we assume herein that each agent holds exactly one variable.} 
$\cD = \{D_1, D_2, \ldots , D_n\}$ 
is a set of domains, in which each domain $D_i$ consists of the finite set of values that can be assigned to variable $X_i$; $\cR$ is the set of relations (constraints), where relation $R \in \cR$ is a function $R : D_{i_1} \times D_{i_2} \times \ldots \times D_{i_k} \rightarrow \prod^k_{j = 1} \mathbb{R}_\geq$ that defines a non-negative utility for every participant in every value combination of a set of variables. The asymmetry of relations in the ADCOP model stems from the potentially different utilities for every participant. 

An assignment is a pair including a variable and a value from that variable's domain. A solution is a set of assignments for all the variables in $\cX$. An optimal solution is a solution of aggregated maximal utility.\footnote{Commonly (A)DCOP constraints impose costs rather than utilities, i.e., the optimal solution is aggregated minimal cost. Herein, we use utilities; hence the objective is maximization.} 

The inherent distribution of (A)DCOP has several important advantages. First, it eliminates the need for a human coordinator. Another important aspect is that of privacy -- in a distributed environment, unlike a centralized one, agents need not a-priori reveal their entire preferences to some central entity that solves the problem. Subsequently, a distributed environment has no single point of failure.

\section{Course allocation with friendships}  
\label{sec:solution}

In this section, we present our proposed solution. We start by describing in Subsection~\ref{sec:model} our course allocation model that considers friendships. Next, in Subsection~\ref{sec:ADCOPformulation}, we formulate the problem as an ADCOP. Then, in Subsection~\ref{sec:algorithm}, we introduce a novel algorithm that utilizes the specific features of the problem at hand. Finally, we discuss some implementation issues in Subsection~\ref{sec:implementation}. 

\subsection{Model}\label{sec:model}
Our model consists of $n$ students ($s_1,\ldots,s_n$) that require allocation to courses from a set $C$ of available courses. We denote the size of this set $m=|C|$. Each course $c\in C$ has a capacity $q_c$, corresponding to the number of available seats in that course. We follow the common assumption in course allocation literature that all courses have the same capacity. We denote that capacity by $q$.
Each student should be allocated to several courses. 
A bundle set $B_i \subset C$ is the course allocation for student $s_i$. 
We follow~\citet{budish2017course} and assume that there is a common bundle size $b=|B_1|=\cdots =|B_n|$ for all students.


The course utility is a function that represents the prospects the students gain from their assigned bundles. 

\begin{definition}[Course Utility]
Let $r_{i}(c)$ denote the reward (utility) of student $s_i$ from course $c$. Consequently, the student's course utility from a bundle $B_i$ is:

\begin{equation}\label{eq:unary_eq}
u_p(i) := \sum_{c \in B_i}{r_{i}(c)}
\end{equation}
\end{definition}

In this work, we consider an additional utility --- the utility students gain from studying with their friends in the same course. 

\begin{definition}[Friendship Utility]
The utility that student $s_i$ gains from being allocated to course $c$ with student $s_j$ is:
$ r_{i}(s_j,c)\cdot w$. 
The tuning parameter $w$ controls the weight given to friendships. When $w=0$, friendships have no utility. 
The friendship utility of student $s_i$ is the sum of utilities from all of their friends who share courses with them: 
\begin{equation}\label{eq:binary_eq}
    u_f(i) := \sum_{c \in B_i} \sum_{j|c\in B_j}{r_{i}(s_j,c)\cdot w}
\end{equation}
\end{definition}

Note that the friendship utility is non-zero only when $s_i$ and $s_j$ are in the same course and that the utility may be asymmetric, i.e., 
$ r_{i}(s_j,c)\cdot w \neq r_{j}(s_i,c)\cdot w$.

A feasible allocation is such that: (a) for each course $c$, there are no more than $q_c$ allocated students, and (b) each student $s_i$ receives an allocation bundle $B_i$ of size $b$. 
The objective is to find a feasible allocation with maximal total utility.

\begin{definition}[Total Utility]
The total utility is:
\begin{equation} \label{eq:utility}
    \sum_{i=1}^{n} (u_p(i) + u_f(i))
\end{equation}
\end{definition}






\subsection{ADCOP formulation}\label{sec:ADCOPformulation} 

We now formulate the proposed model as an ADCOP. Recall (Section~\ref{subsec:adcop}) that ADCOP is a tuple 
$<\cA,\cX,\cD,\cR>$. 
In our formulation, the agents $\cA$ are the $n$ students. 
The variables $\cX$ that the agents hold represent the bundle of courses. The domains $\cD$ are the sets of all possible bundles, i.e., 
$\binom{m}{b}$.
We consider three types of relations (constraints) in $\cR$: 
\begin{itemize}
    \item Unary constraints -- each student has a unary table (vector) that indicates their course preferences.
    \item Binary constraints -- for every two students, we create a binary table containing a Cartesian product of all possible value assignments in the domain. Each cell in the table receives a value based on the number of common courses among the two students multiplied by the students' friends rating. Note that the binary table is asymmetric, which means that for every two agents, there are two different tables.
    \item Global constraints -- hard constraints that enforce the course capacity limit $q$.
\end{itemize}

To better understand the formation of the unary and binary constraints, consider the following example.

Consider $n=3$ students Alice, Bob, and Charlie; $m=4$ available courses $c_1$, $c_2$, $c_3$, and $c_4$; and a required course bundle of size $b=3$.
Before attending to the generation of constraints, we first view the students' preferences over courses (utility $u_p$) and their friendship ratings (utility $u_f$, $w=2$). These are found in Tables~\ref{table:stu_prefs} and~\ref{table:fr_prefs}, respectively.
Since it is a small example, all students are constrained by each other; however, in general, not all students must be constrained by each other.

\begin{table}[h!]
\centering
\caption{Students' preferences over courses}
\begin{tabular}{|c c c c c|}  
 \hline
   & $c_1$ & $c_2$ & $c_3$ & $c_4$ \\
 \hline
 Alice  & 4 & 3 & 2 & 1 \\
 \hline
 Bob  & 3 & 1 & 4 & 2 \\
 \hline
 Charlie  & 1 & 2 & 3 & 4 \\
\hline
\end{tabular}
\label{table:stu_prefs}
\end{table}

\begin{table}[h!]
\centering
\caption{Friendship preferences over students}
\begin{tabular}{|c c c c|} 
 \hline
  & Alice & Bob & Charlie \\
 \hline
 Alice  & X & 6 & 4 \\
 \hline
 Bob  & 2 & X & 6 \\
 \hline
 Charlie & 4 & 2 & X \\
\hline
\end{tabular}
\label{table:fr_prefs}
\end{table}

Now we turn to the formation of the constraints. We start with the unary constraints. The size of all the domains is 4 because $m=4 \choose b=3$. Table~\ref{table:unary} shows Alice's unary constraint. 

\begin{table}[h!]
\centering
\caption{Alice's unary constraint}
\begin{tabular}{|c c c c c|} 
 \hline
   & $\{c_1$, $c_2$, $c_3\}$ & $\{c_1$, $c_2$, $c_4\}$ & $\{c_1$, $c_3$, $c_4\}$ & $\{c_2$, $c_3$, $c_4\}$ \\
 \hline
 Alice & 9 & 8 & 7 & 6 \\
 \hline
\end{tabular}
\label{table:unary}
\end{table}

Next, we present the binary constraints. Each binary constraint is a table of size 4X4. Table~\ref{table:binary} displays the relation between Alice and Bob. Recall that each student holds a separate binary constraint due to the asymmetry of friendships. Nevertheless, for ease of presentation, we use a single table with a pair (a~,~b) in each cell, where 'a' represents Alice's binary constraint and 'b' represents Bob's constraint.

\begin{table}[h!]
\centering
\caption{Binary constraints of Alice and Bob}
\begin{tabular}{|c c c c c|} 
 \hline
   & $\{c_1$, $c_2$, $c_3\}$ & $\{c_1$, $c_2$, $c_4\}$ & $\{c_1$, $c_3$, $c_4\}$ & $\{c_2$, $c_3$, $c_4\}$ \\
 \hline
 $\{c_1$, $c_2$, $c_3\}$ & (18~,~6) & (12~,~4) & (12~,~4) & (12~,~4) \\
 \hline
 $\{c_1$, $c_2$, $c_4\}$ & (12~,~4) & (18~,~6) & (12~,~4) & (12~,~4) \\
 \hline
 $\{c_1$, $c_3$, $c_4\}$  & (12~,~4) & (12~,~4) & (18~,~6) & (12~,~4) \\
\hline
 $\{c_2$, $c_3$, $c_4\}$  & (12~,~4) & (12~,~4) & (12~,~4) & (18~,~6) \\
\hline
\end{tabular}
\label{table:binary}
\end{table}

Assume that Alice and Bob were both allocated the course bundle: $c_1$,$c_2$,$c_3$. Charlie was allocated the bundle: $c_2$,$c_3$,$c_4$. Their utilities according to Equations (\ref{eq:unary_eq}) -- (\ref{eq:utility}) are as follows:

\begin{itemize}
    \item Alice's Utility: 
    \\ $u_p(Alice) = 4 + 3 + 2 = 9$ 
    \\ $u_f(Alice) = 3 \cdot 6 + 2 \cdot 4 = 26$ 
    \\ $u_p(Alice) + u_f(Alice)  = 35$. 
    \item Bob's Utility: 
    \\ $u_p(Bob) = 3 + 1 + 4 = 8$ 
    \\ $u_f(Bob) =  3 \cdot 2 + 2 \cdot 6 = 18$ 
    \\ $u_p(Bob) + u_f(Bob)  = 26$. 
    \item Charlie's Utility: 
    \\ $u_p(Charlie) = 2 + 3 + 4 = 9$ 
    \\ $u_f(Charlie) = 2 \cdot 4 + 2 \cdot 2 = 12$ 
    \\ $u_p(Charlie) + u_f(Charlie)  = 21$. 
\end{itemize}

\subsection{Algorithm}\label{sec:algorithm} 

Our formulation requires all agents to communicate with each other for two reasons: (a) there is a global constraint for enforcing the course capacity $q$; (b) the friendship relations are asymmetric, thus an agent may not be aware which other agents are constrained with it (i.e., consider it as their friend). This forms a communication clique. Such a clique, together with the typically large number of students in course allocation problems, yields the use of an incomplete lightweight algorithm.   
Thus, we first introduce the classical Distributed Stochastic Search Algorithm (\dsa)~\citep{zhang2005distributed}, which is a simple, yet very successful, local search algorithm for standard (symmetric) DCOP.

\begin{algorithm}
\SetAlgoLined
\SetKwFunction{algo}{\textsc{DSA}}
  \SetKwProg{myalg}{Algorithm}{}{}
  \myalg{\algo{$\alpha, rounds$}}{
$counter$ $\gets$ 0\;
$value$ $\gets$ Random(domain)\;
send $value$ to neighbors\;
\While{$counter < rounds$}{
collect neighbors' values\;
\If {$\textsf{\upshape Random([0..1)})$ $< \alpha$}{
$value$ $\gets$ value with maximal utility\;
send $value$ to neighbors\;}
$counter$ $\gets$ $counter+1$\; 
}
}
\caption{\dsa algorithm}
\label{alg:dsa}
\end{algorithm}

The pseudo-code of the standard \dsa algorithm is given in Algorithm~\ref{alg:dsa}. It begins with each agent randomly assigning a value to its variable (line 3) and sending that value to all neighbors, i.e., agents with whom there exists a constraint (line 4).
The agents run the algorithm for a predetermined number of rounds, which is received as input (lines 1,5). In each round, an agent collects the assignments sent by its neighbors (line 6). 
Then, each agent faces a decision whether to keep its current value assignment or to try to improve it. The decision is made stochastically according to a predetermined probability $\alpha$ (lines 1,7).\footnote{The predefined probability is often denoted $p$. However, we denote it $\alpha$ herein for better compatibility with the new algorithm that we propose later in this section.} The use of probability $\alpha$ prevents the possibility of infinite cycles in which all agents change their assignments simultaneously. In case the decision is of a change, the agent tries to improve its current state by replacing its current value in a manner that maximizes its utility (line 8). No change is made if the agent cannot improve its current state. New value assignments are sent to all neighbors (line 9), which collect them in the next round (line 6).

The standard \dsa does not consider the constraints that stem from the asymmetry, i.e., the constraints perceived by the peer agents. The Asymmetric Coordinated Local Search (\acls) algorithm was proposed for that purpose~\citep{grinshpoun2013asymmetric}. \acls coordinates between the agents sharing a constraint in order to reach more efficient and stable solutions. However, \acls considers a slightly different notion of asymmetry, as it assumes that the asymmetry only regards the costs/utilities of the constraint's sides. Contrary to that, our model assumes asymmetry in a broader sense, in which the mere existence of a constraint is asymmetric. As a consequence, the coordination mechanism of \acls reveals vital private information regarding friendships when applied to our model.

Our model yields an additional problem to general-purpose algorithms, such as \dsa and \acls, which relates to the course capacity constraints. Such constraints have two important features -- they are \textit{global}, i.e., all the agents participate in these constraints, and they are \textit{hard}, i.e., any valid solution must adhere to these constraints. These features drove us to develop a novel algorithm, \dsarc, which can explicitly handle such \textit{resource capacity} constraints.

\begin{algorithm}
\SetAlgoLined
\SetKwFunction{algo}{\textsc{DSA\_RC}}\SetKwFunction{proc}{MinConflictMaxUtil}
  \SetKwProg{myalg}{Algorithm}{}{}
  \myalg{\algo{$\alpha, rounds, q$}}{
$counter \gets$ 0 \;
$value \gets$ Random(domain) \;
send $value$ to all agents \;
\While{$counter <  rounds$}{
collect agents' values \;

$invalid\_courses$ $\gets$ courses in $value$ that exceed $q$ \;
\If {$invalid\_courses \neq \emptyset$}{
$max\_course$ $\gets$ course in $invalid\_courses$ with maximal deviation beyond capacity $q$ \;
$max\_students$ $\gets$ number of assigned students in $max\_course$ \;
$\beta \gets \frac{max\_students - q}{max\_students}$ \;
\If {$\textsf{\upshape Random([0..1)})$ $< \beta$}{
$value$ $\gets$ valid value with maximal utility  \;
\If{$value = \emptyset$}{
$value$ $\gets$ MinConflictMaxUtil() \;}
send $value$ to all agents \;}}
\ElseIf {$\textsf{\upshape Random([0..1)})$ $< \alpha$}{
$value$ $\gets$ valid value with maximal utility  \;
send $value$ to all agents \;}
$counter$ $\gets$ $counter+1$\;} 
}
\setcounter{AlgoLine}{0}
  \SetKwProg{myproc}{Procedure}{}{}
  \myproc{\proc{}}{
$valid\_courses$ $\gets$ courses with available seats ($<q$) \;
generate $temp\_domain \subset$ domain, in which all values contain all $valid\_courses$ \;
$value$ $\gets$ value with maximal utility in $temp\_domain$ \;
\textbf{return} $value$ \;
}
\caption{\dsarc algorithm}
\label{alg:dsarc}
\end{algorithm}

The pseudo-code of the \dsarc algorithm is given in Algorithm~\ref{alg:dsarc}. The algorithm begins (lines 1-6) and ends (lines 17-20) similarly to \dsa, with two differences: first, now values are sent to all the agents and not just to neighbors because of the asymmetry of friendships and the global capacity constraints; second, the validity of new values must be confirmed due to the hard capacity constraints (see line 18 in Algorithm~\ref{alg:dsarc} vs. line 8 in Algorithm~\ref{alg:dsa}). 

The novel part of \dsarc deals with resource capacity constraints. In line 7 a list of invalid courses is generated; a course is considered invalid if it is one of the courses within the current value assignment and the seat capacity $q$ is exceeded according to the assignments of the other students. It becomes interesting when the list of invalid courses is not empty (lines 8-16), otherwise, \dsarc continues as \dsa (lines 17-20).

In case of multiple invalid courses, the course with the largest overflow beyond the seating capacity is chosen as $max\_course$ (line 9). The variable $max\_students$ corresponds to the number of students in that course (line 10). Next, the $\beta$ parameter is computed by dividing the overflow in $max\_course$ ($max\_students - q$) by $max\_students$ (line 11). The rationale behind this parameter is to compute the needed fraction of agents that should remove $max\_course$ from their value assignment. Indeed, this is exactly the probabilistic condition applied in line 12. If the condition holds (lines 13-16), a new $valid$ value assignment is chosen for the agent, i.e., excluding courses with no available seats (line 13). The new value assignment is then sent to all the agents (line 16).

In some extreme cases, there might be no valid solution because the number of courses with remaining available seats is smaller than the course bundle size $b$ the student must attend. These cases are handled in line 15 by a call to procedure MinConflictMaxUtil(). As its name suggests, the procedure first minimizes the number of conflicts (i.e., courses with overflow) and then chooses the bundle with maximal utility out of the remaining options. The procedure starts with finding the list of valid courses, i.e., courses with currently available seats. Next, in line 3 of the procedure, a temporary domain is constructed of all course bundles of size $b$ that contain all the valid courses. The valid courses are complemented with invalid course/s to reach bundle size $b$. This forms the set of values with a minimal number of conflicts. Finally, from this temporary domain, the value with maximal utility is chosen (line 4) and returned (line 5).


\subsection{Implementation issues}\label{sec:implementation} 

We implemented \dsarc on the AgentZero simulator\footnote{AgentZero Tutorial, including installation instructions: \url{https://docs.google.com/document/d/1B19TNQd8TaoAQVX6njo5v9uR3DBRPmFLhZuK0H9Wiks/view}}~\citep{lutati2014agentzero}. The simulator enables to remember the best solution obtained in any stage of the solution process. Such a feature turns \dsarc to an \textit{anytime} algorithm~\citep{zilberstein1996using}, i.e., an algorithm that ensures that the obtained solution only improves over time. 
\dsarc can be adjusted to be an anytime algorithm in truly distributed environments by applying the method of \citet{zivan2014explorative}.

Another implementation issue regards the global course capacity constraints. Instead of modeling cumbersome global constraints, we hard-wired the needed computation into the \dsarc code. This coincides with the explicit checks of validity in the pseudo-code (e.g., lines 9, 13, 18).

Our source code is available online: 

\url{https://github.com/Justrygh/Course-allocation-with-friends}.

\section{Evaluation} 
\label{sec:eval}
We first present the experimental setups, followed by our results. 

\subsection{Experimental setup} 
As the problem is NP-hard, exact methods, such as linear programming algorithms, are inapplicable for realistic settings with tens to hundreds of students. Therefore, we follow the state-of-the-art and focus on scalable solutions.
We evaluated the following algorithms: 
\begin{itemize}
    \item \dsarc~-- our main algorithm, presented in Section \ref{sec:algorithm}. We set the number of rounds to $50$ and $\alpha = 0.8$.  
    \item \dsa~-- presented in Section \ref{sec:algorithm}. Again, we set the number of rounds to $50$ and $\alpha = 0.8$.  
    \item \hbs~-- presented in Section \ref{subsec:courses}. We adjusted \hbs to consider utilities from friendships when the current student's friends have already selected courses.
    \item \rsd~-- short for Random Serial Dictatorship. An iterative greedy algorithm that is used as a benchmark~\citep{budish2012multi}. Students are randomly ordered. In turn, they choose from valid assignments (i.e., courses with available seats) unless no such assignments exist. 
    We adjusted \rsd to consider utilities from friendships.
    \item \greedy~-- each student selects their favorite $b$ courses. All selections are performed in parallel. 
    \item \random~-- each student selects $b$ \textit{random} courses. All selections are performed in parallel.   
\end{itemize}

In order to evaluate how the algorithms fare in terms of social welfare, fairness in order, and equality, we used the following evaluation metrics: 

\begin{enumerate}
    \item Courses utility -- we assume that the students have strict preferences over the courses: $r_i(c) \succ r_i(c') \succ \cdots \succ r_i(c'') $. 
    The utility of the first-ranked course is $u(r_i(c)) = m$; the utility of the second-ranked course is $u(r_i(c')) = m-1$; and the utility continues to decrement until the minimal utility $u(r_i(c'')) = 1$ is given to the last-ranked course.
   The utility is then calculated according to Equation \ref{eq:unary_eq}.
    \item Friendship utility -- as with the course utility, we set the friendship utility according to the ranking of the friends. The utility for the top-most, second-, and third-ranked friends is $3\cdot w$, $2\cdot w$, and $1\cdot w$, respectively. The utility is then calculated according to Equation \ref{eq:binary_eq}.
    \item Total utility -- calculated according to Equation \ref{eq:utility}.
    \item Number of illegal assignments -- indicates the allocation's validity.     
    \item First/Middle/Last Agents' utility -- the average utility for the first, middle, and last agents. This metric allows us to obtain insights regarding the fairness of the agents' order. 
    \item Gini coefficient -- indicates the equality fairness among the agents \citep{gini1936measure}. 
    \begin{equation} \label{eq:gini}
        G = \frac{\sum_{i=1}^{n} \sum_{j=1}^{n} | (u_p(i) + u_f(i)) - (u_p(j) + u_f(j)) |}{2 \sum_{i=1}^{n} \sum_{j=1}^{n} u_p(j) + u_f(j)} 
    \end{equation}
\end{enumerate}

The social welfare is examined in the first three metrics.
The general validity of the solution is examined via the fourth metric. Fairness in order is the purpose of the fifth metric, and equality is analyzed in the sixth metric. 
Note that all of the evaluated heuristics are lightweight and thus fast. For example, running a \dsarc instance took 3-4 seconds on hardware with an Intel i5-7200U processor and 8GB memory. Therefore, we did not include a run-time evaluation.

We set the parameters to correspond with real-world elective graduate courses at our university:
\begin{itemize}
    \item Number of courses available: $m=9$. 
    \item Bundle size: $b=3$
    \item Number of students (agents): $n$. This number varied for the simulations and was set to $n=177$ in the user study.
    \item Course capacity limit: $q$. This number was set to $q=30$ in the simulations and varied in the user study: $q \in \{60,65,\ldots,90\}$.
    \item Course bundle size, $b$. The number of courses each student is required to take. As mentioned in Section \ref{sec:model}, we assume that all students require the same bundle size. We set $b = 3$.
    \item Number of friends, $f$.  To avoid unfair situations where students with more friends gain a higher utility, we set an equal number of $f=3$ friends for all agents.  
    \item Friendship weight, $w$. First, we examine the algorithms with a friendship weight $w = 2$. Afterward, we present the algorithms with different friendship weights $w \in \{0,...,5\}$.   
\end{itemize}



\subsection{Results}
We examined simulated data, which is based on real-world preferences. We also examined data we collected in a user study. 

\noindent \textbf{Simulated Data:}         
    To set the students' preferences over courses, we used the Courses 2003 dataset from the Preflib library~\citep{MaWa13a}. The dataset contains 146 students and their strict preferences over nine courses. We randomly assigned three friends for each student.
    We sampled with return various student preferences: $n \in \{40,50,\ldots,90\}$. We set the course capacity limit to $q = 30$.
    For each sample of $n$, we randomly generated 50 problems; each data point in the presented graphs represents the average of these 50 runs. 

\noindent \textbf{User Study:}
    We conducted an experiment within the Industrial Engineering and Management Department at Ariel University, Israel. Our study received ethical approval from the institutional review board of Ariel University.
    Participants were recruited from second-year students. In total, 184 second-year students agreed to participate. They received a list of nine elective courses offered to third-year students. Participants were asked to rank the courses according to their preferences. Additionally, they were asked to select and list in descending order three students they would like to study the courses with. Out of 184 participants, seven students failed to report all of their preferences and were removed from the experiment. Therefore, 177 valid participants $n = 177$ were included in the experiment. 
    We varied the course capacity limit to $q \in \{60,65,\ldots,90\}$.
    For each setup, we randomly generated 50 problems such that the only thing that changed between the problems was the order of the students; each data point in each of the presented graphs represents the average of these 50 runs. 

\begin{figure} 
\includegraphics[scale=0.45]{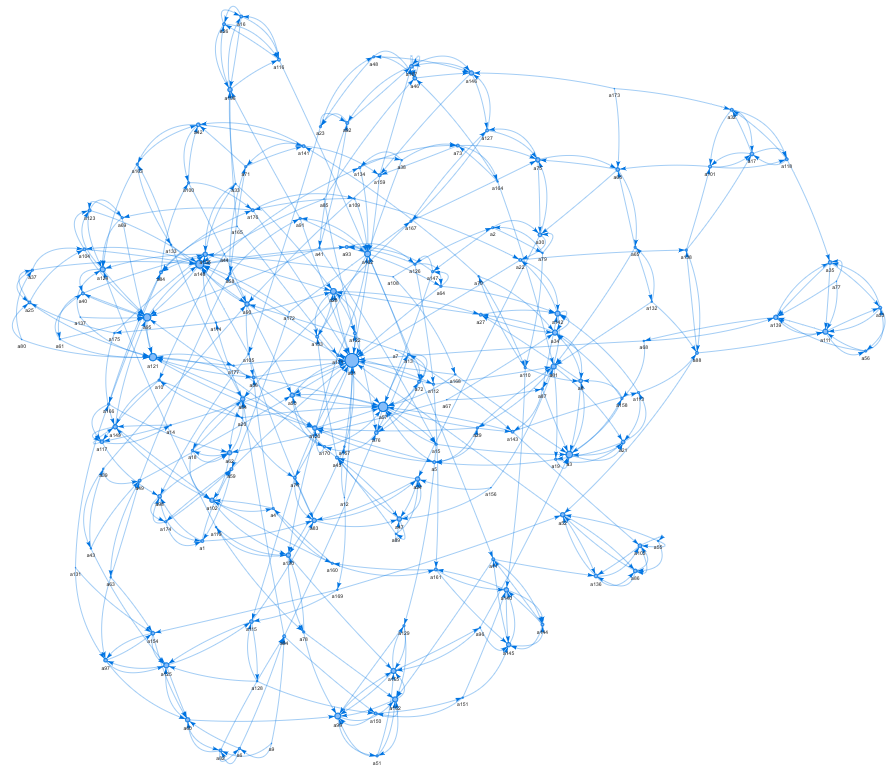}
\caption{The friendship network defined in the problem (user study)}
\label{fig:friendship_network}
\end{figure}

Figure~\ref{fig:friendship_network} presents the friendship network as a directed graph $s.t.$ each vertex is an agent, and a directed arc from $A$ to $B$ indicates that agent $A$ considers agent $B$ a friend. The number of incoming arcs determines the size of a vertex.

\begin{figure}     
    \includegraphics[scale=0.7]{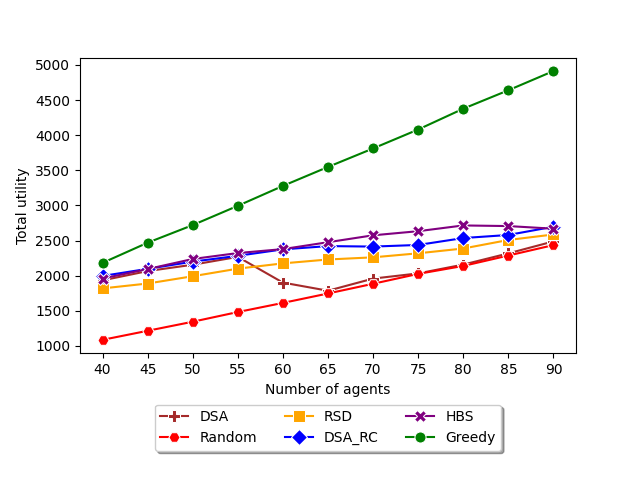}
    \caption{The total utility as a function of the number of students (simulated data)}
    \label{fig:sim_total_utility}    
    \includegraphics[scale=0.7]{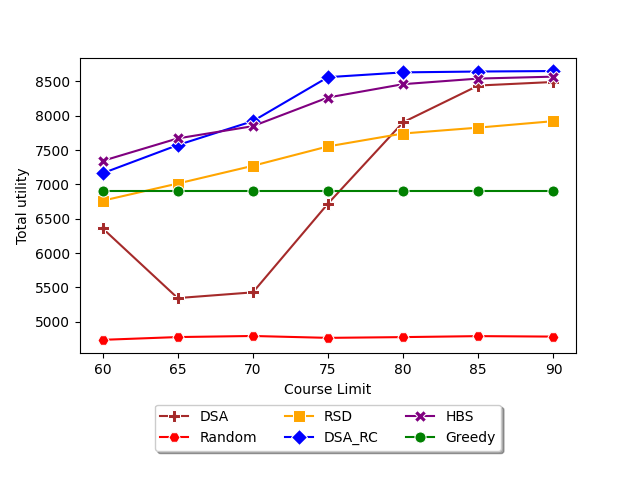}
    \captionof{figure}{The total utility as a function of the course limit  (user study)}
    \label{fig:study_total_utility}
\end{figure}

Figure~\ref{fig:sim_total_utility} presents the utility of the six algorithms using simulated data. 
The y-axis displays the total utility of the students, and the x-axis depicts the number of students. 
As expected, \greedy has the highest utility since each agent chooses an assignment based on its utility without considering seat capacity. Also, as expected, \random has the lowest utility since the assignments are chosen randomly without considering the agent's utilities from courses or friends. 
\dsarc, \hbs, and \rsd outperform \dsa.
The total utility achieved by \hbs is slightly higher than that of \dsarc when the number of agents is between 70 to 85, i.e., when the problem is more constrained.   

Figure~\ref{fig:study_total_utility} presents the utility in the user study data. In the user study, the number of students is constant, so we used the x-axis to examine various course capacity limits. 
\greedy has a constant utility since each agent chooses an assignment based on its utility without considering the capacity. 
Again, \random is provided as a baseline and has the lowest utility.
\dsarc and \hbs obtain the best performance. 
The total utility achieved by \dsarc is slightly higher when the course limit exceeds 65 seats and the problem is less constrained. 

\begin{figure}     
    \includegraphics[scale=0.7]{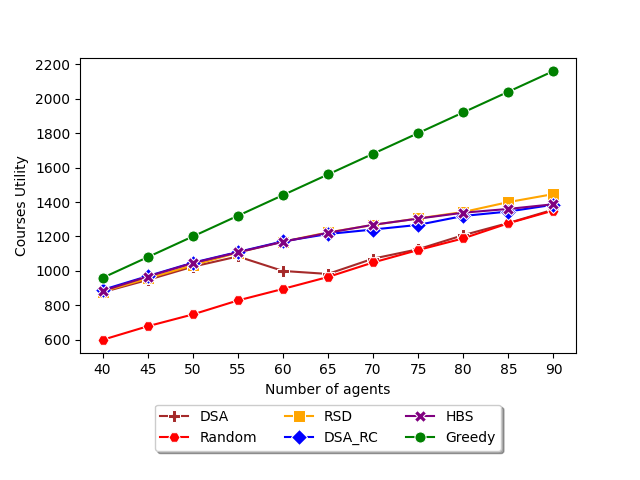}
    \caption{The course utility as a function of the number of students (simulated data)}
    \label{fig:sim_course_utility}    
    \includegraphics[scale=0.7]{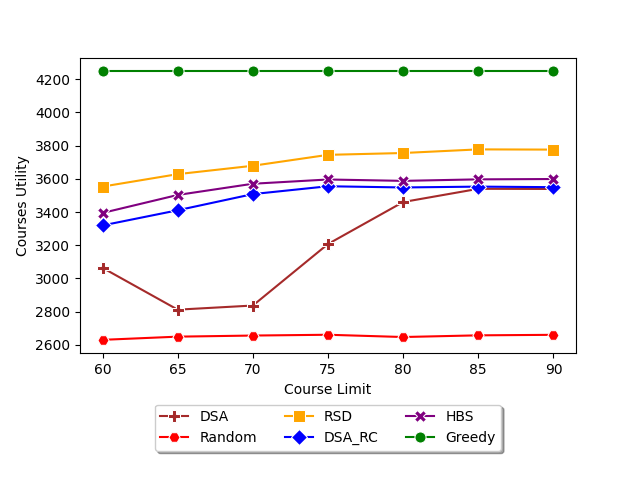}
    \captionof{figure}{The course utility as a function of the course limit  (user study)}
    \label{fig:study_course_utility}
\end{figure}

The total utility consists of course utility (the utility students gain from their assigned courses) and friendship utility (the utility students gain from sharing courses with their friends). We now examine each of these utilities separately. 
Figures~\ref{fig:sim_course_utility} and~\ref{fig:study_course_utility} present the course utility in the simulated data and user study, respectively. 
In the simulated data, The course utility achieved by \hbs, \dsarc, and \rsd is almost the same, while \dsa has a lower utility.
In the user study, apart from \greedy, \rsd has the highest course utility, followed by \hbs and \dsarc.

\begin{figure}     
    \includegraphics[scale=0.7]{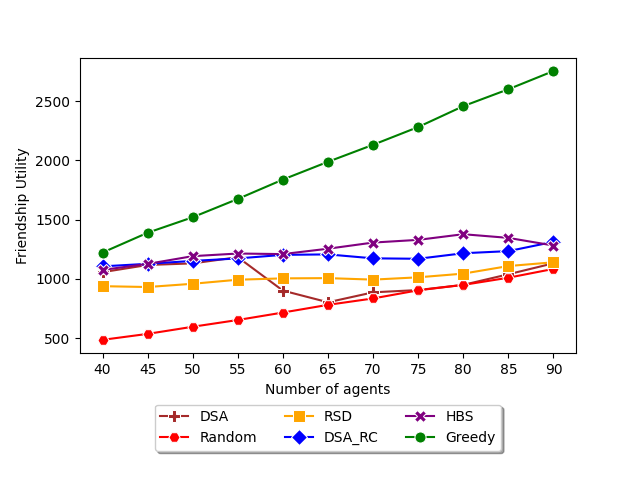}
    \caption{The friendship utility as a function of the number of students (simulated data)}
    \label{fig:sim_friend_utility}    
    \includegraphics[scale=0.7]{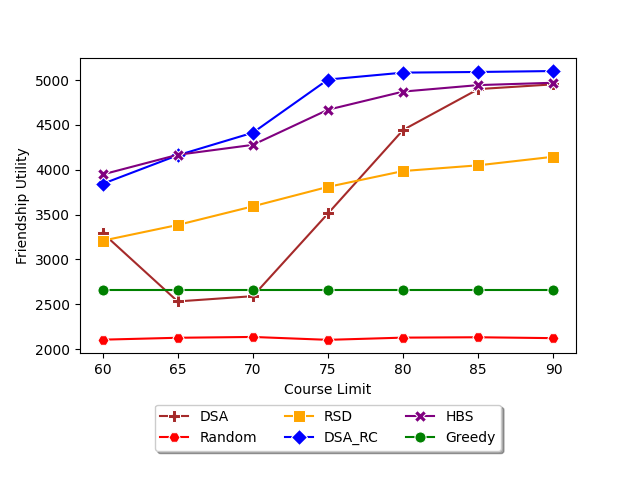}
    \captionof{figure}{The friendship utility as a function of the course limit  (user study)}
    \label{fig:study_friend_utility}
\end{figure}

Figure~\ref{fig:sim_friend_utility} and~\ref{fig:study_friend_utility} present the friendship utility in the simulated data and user study, respectively. 
In the simulated data, \rsd, \dsarc, and \hbs exhibit similar trends to that in Figure~\ref{fig:sim_total_utility}. 
Although \greedy and \random do not explicitly consider friendships, their utilities increase with the number of students. This phenomenon will be explained in the next set of figures. 
In the user study, \dsarc outperforms \hbs in all but the most constrained setting (when the course capacity limit is 60). 
\greedy maintains a constant utility, as the friendship utility is unaffected by the course limit.

\begin{figure}  
    \includegraphics[scale=0.7]{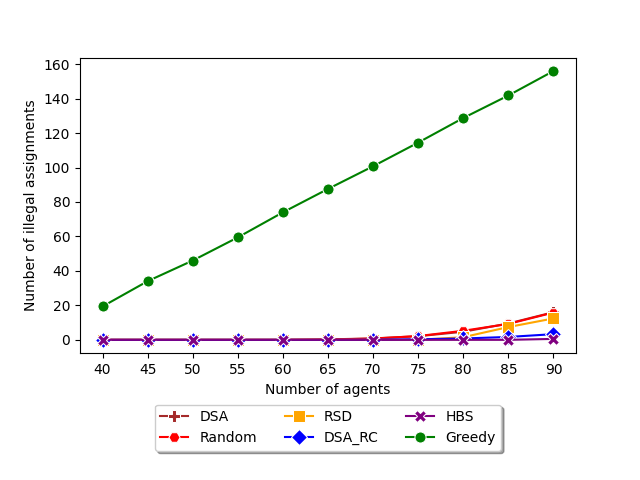}
    \caption{The number of illegal assignments as a function of the number of students (simulated data)}
    \label{fig:courses}    
    \includegraphics[scale=0.7]{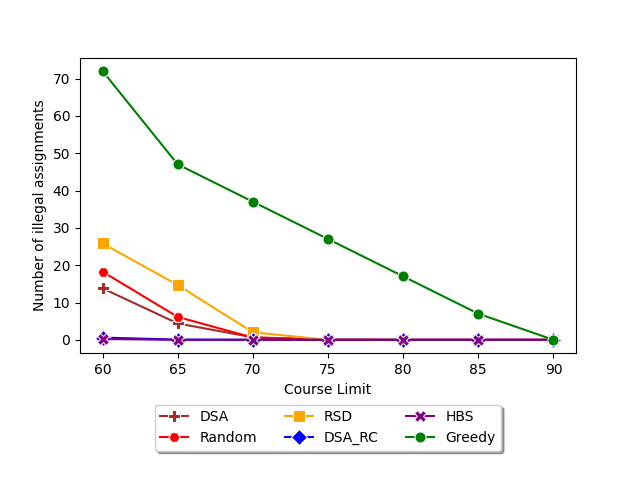}
    \captionof{figure}{The number of illegal assignments as a function of the course limit  (user study)}
    \label{fig:courses_x2}
\end{figure}

We next evaluated the number of illegal assignments, i.e., the allocation feasibility as displayed in the number of students that exceed the course limit in each course.
Figures~\ref{fig:courses} and~\ref{fig:courses_x2} present the number of illegal assignments of the six algorithms, as a function of the number of students (Figure~\ref{fig:courses}, simulated data) and as a function of the course limit (Figure~\ref{fig:courses_x2}, user study). 
Here, \greedy has the highest number of illegal assignments since each agent chooses its best assignment without considering the problem constraints. 
This explains why the friendship utility of \greedy grows linearly in the number of agents (Figure~\ref{fig:sim_friend_utility}):  
a high friendship utility comes at the cost of an increased number of illegal assignments. 
The other algorithms find valid solutions when the setting is not significantly constrained (40-70 agents or 75-90 course seats). However, only \dsarc and \hbs can find valid solutions in constrained settings. 

\begin{figure} 
\centering
\includegraphics[scale=0.5]{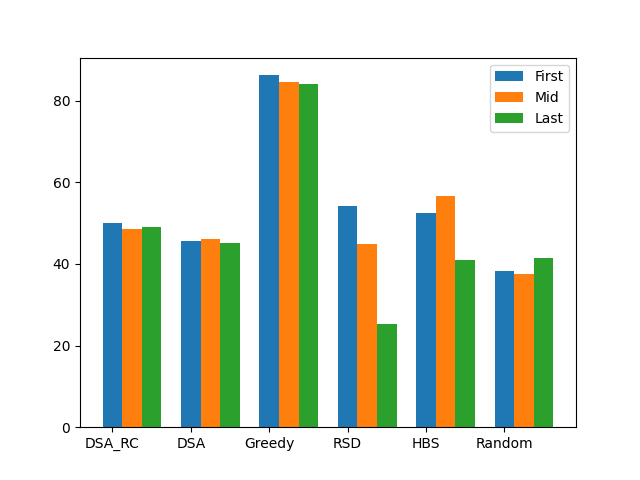}
\caption{The utility of the first, middle, and last agents for problems with 80 students (simulated data)}
\label{fig:utility_x2_agents_80}
\includegraphics[scale=0.5]{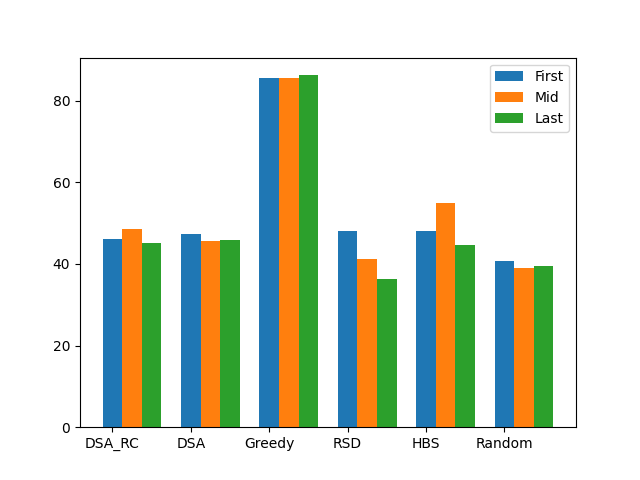}
\caption{The utility of the first, middle, and last agents for problems with 85 students (simulated data)}
\label{fig:utility_x2_agents_85}
\includegraphics[scale=0.5]{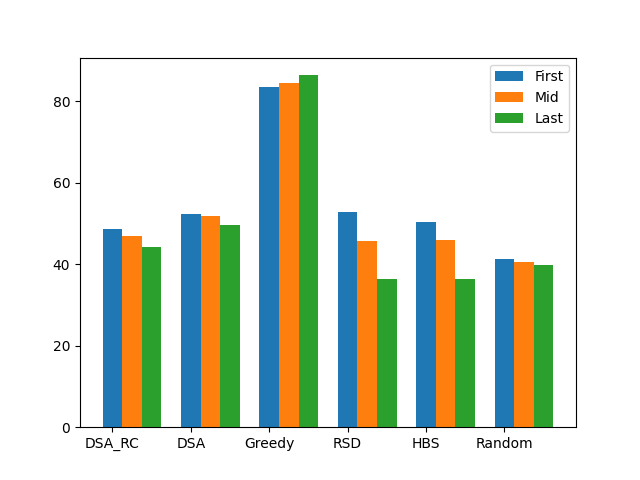}
\caption{The utility of the first, middle, and last agents for problems with 90 students (simulated data)}
\label{fig:utility_x2_agents_90}
\end{figure}

\begin{figure} 
\centering
\includegraphics[scale=0.5]{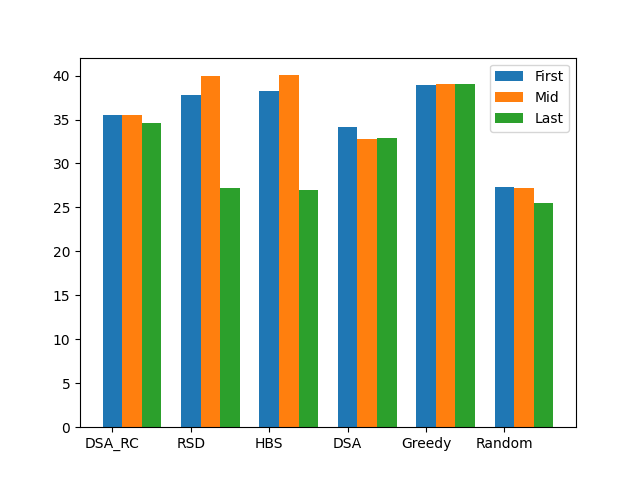}
\caption{The utility of the first, middle, and last agents given course capacity of 60 students (user study)}
\label{fig:utility_x2_courseLimit_60}
\includegraphics[scale=0.5]{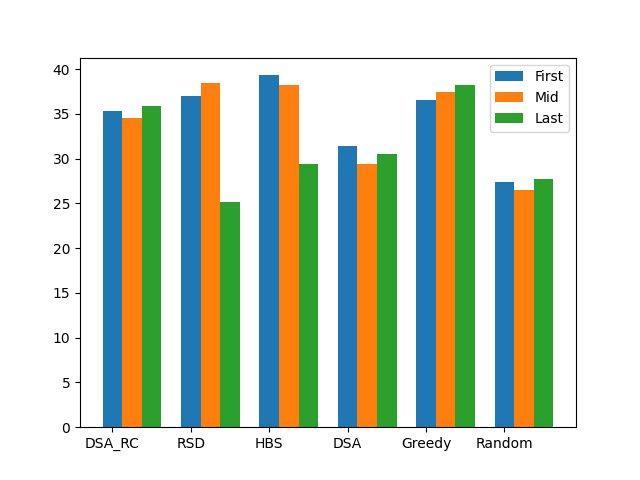}
\caption{The utility of the first, middle, and last agents given course capacity of 65 students (user study)}
\label{fig:utility_x2_courseLimit_65}
\includegraphics[scale=0.5]{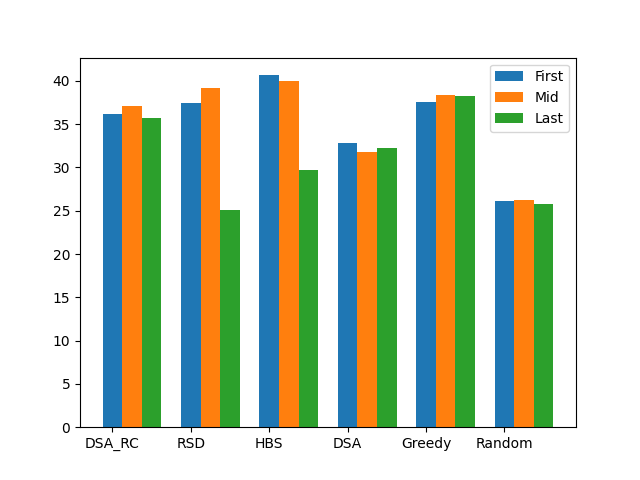}
\caption{The utility of the first, middle, and last agents given course capacity of 70 students (user study)}
\label{fig:utility_x2_courseLimit_70}
\end{figure}

We then examined the fairness in order. Figures~\ref{fig:utility_x2_agents_80},~\ref{fig:utility_x2_agents_85}, and~\ref{fig:utility_x2_agents_90} display the utility of the first, middle, and last agent in the simulated data when the number of agents is 80, 85, and 90, respectively.
Similarly, Figures~\ref{fig:utility_x2_courseLimit_60},~\ref{fig:utility_x2_courseLimit_65}, and~\ref{fig:utility_x2_courseLimit_70} display the utilities in the user study when the course capacity limit is 60, 65, and 70, respectively.
For \rsd and \hbs, the first and middle agents enjoy high utility, whereas the last agent's utility is considerably lower. This means that \rsd and \hbs do not maintain fairness in order. In contrast, for \dsarc, \dsa, \greedy, and \random, the utility of the agents is similar regardless of their order (first, middle, or last).

\begin{figure} 
\centering
\includegraphics[scale=0.7]{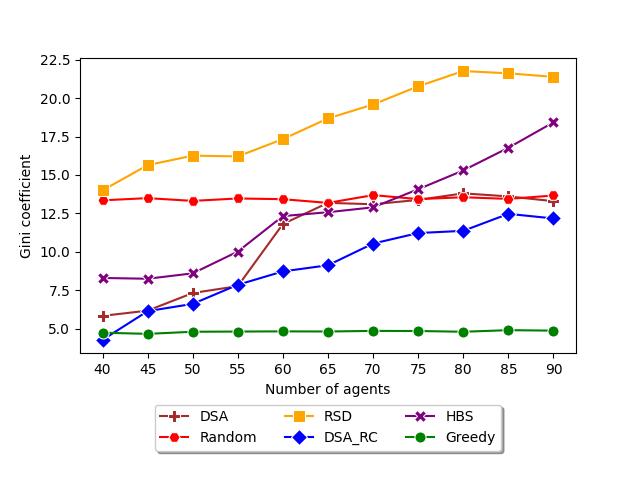}
\caption{The Gini coefficient as a function of the number of students (simulated data)}
\label{fig:gini}
\includegraphics[scale=0.7]{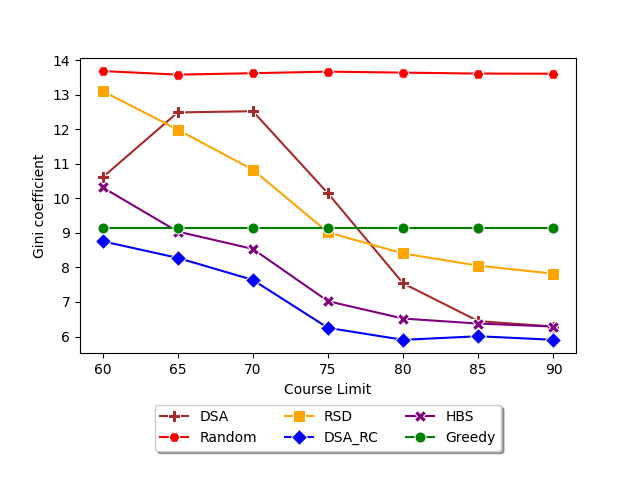}
\caption{The Gini coefficient as a function of the course limit  (user study)}
\label{fig:gini_x2}
\end{figure}

After examining the fairness in order, we turn to look at the overall fairness over all the agents, using the Gini index. This is displayed in  Figures~\ref{fig:gini} and ~\ref{fig:gini_x2}. The y-axis displays the Gini coefficient of the agents, and the x-axis depicts the number of agents and the course capacity limit in the simulated data and user study, respectively. 
A smaller coefficient means that the solution is fairer, i.e., fewer utility differences exist between the agents. The best-performing algorithms here are \dsarc and \greedy.

To summarize our findings, our algorithm, \dsarc, maintains fairness (as expressed by the Gini coefficient) and fairness in order (as expressed by comparing the utilities of the first, middle, and last agents),
almost always finds valid solutions even in very constrained scenarios, and 
provides high utility for the students. 
In very constrained settings, \hbs slightly outperforms \dsarc in the number of illegal assignments and in the utility. However, \hbs does not maintain fairness in order, 
and has a high Gini coefficient, meaning that students might view the solution as unfair since some students will receive their preferred assignments while others will not. 

\subsection{Friendship weights}
An open question is the impact of the importance of friendships on the algorithms. This importance is translated to friendship weights. 
We varied the friendship weights and reran all the experiments. 
We chose to present the results of the user study as it reflects real-world friendships. We present results for the most constrained problems with a course capacity limit of 60;
recall that \hbs displayed some advantage over \dsarc in such highly constrained problems. 

\begin{figure}[H]
    \centering
    \includegraphics[scale=0.6]{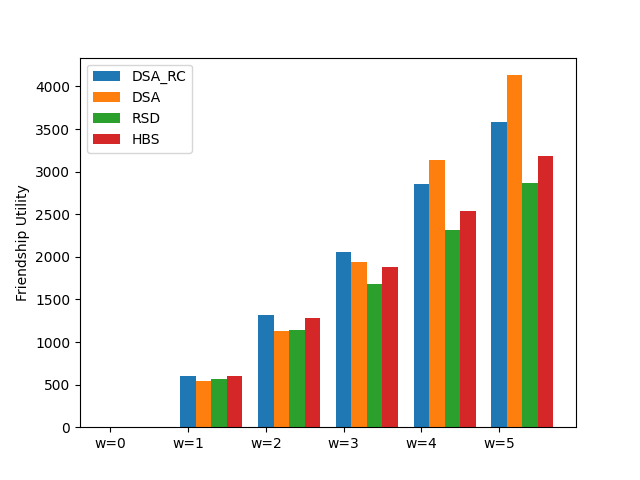}
    \caption{The friendship utility for different friendship weights (user study)}
    \label{fig:experiment_binary_weights}
\end{figure}

Figure \ref{fig:experiment_binary_weights} presents the friendship utility for varying friendship weights.
\dsarc outperforms \hbs in all friendships weights but $w=1$. 
When the friendship weights is $w \in \{4,5\} $, \dsa outperforms \dsarc and has the highest utility.

\begin{figure}[H]
\centering
\includegraphics[scale=0.6]{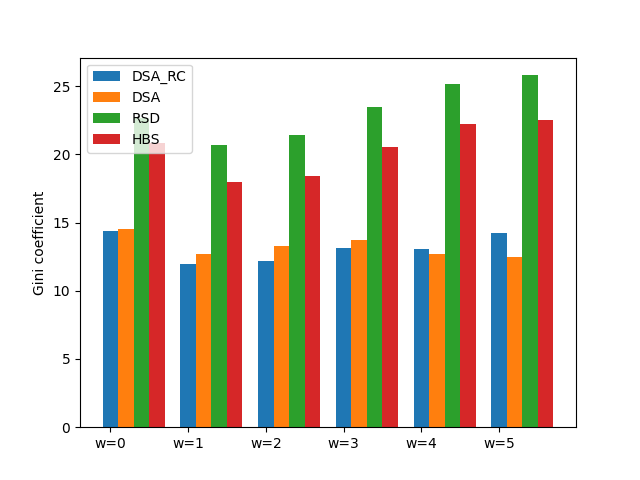}
\caption{The Gini coefficient for different friendship weights (user study)}
\label{fig:experiment_gini_weights}
\end{figure}

Figure~\ref{fig:experiment_gini_weights} presents the Gini coefficient of the four algorithms. A smaller coefficient means that the solution is fairer, i.e., there are fewer differences in utilities between the agents. 
The more the friendships are deemed important (i.e., the more the friendship weight increases), the more \dsarc excels \hbs and \rsd in friendship utility and fairness. For high friendship weights ($w \in \{4, 5\}$), \dsa takes the lead. 

\begin{figure}[H] 
\centering
\includegraphics[scale=0.6]{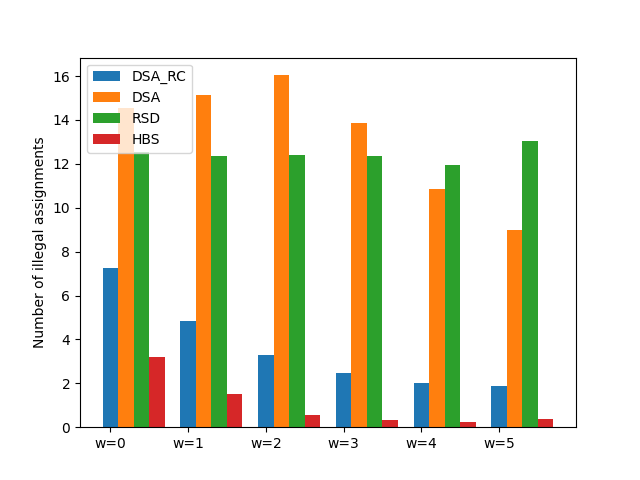}
\caption{The number of illegal assignments for different friendship weights (user study)}
\label{fig:experiment_courses_weights}
\end{figure}

As before, the explanation of this phenomenon lies in the number of illegal assignments, which is presented in Figure~\ref{fig:experiment_courses_weights}.
\dsa has the highest number of illegal assignments, followed by \rsd and \dsarc. In the context of illegal assignments, \hbs remains the leading algorithm even amid different friendship weights.

To conclude, we again find there is a trade-off between utility and fairness and the number of illegal assignments. 

\section{Conclusion} 
\label{sec:conclusion}
In this work, we studied course allocation with friendships. This is a one-sided multi-unit assignment problem. While various solutions to this problem exist, to the best of our knowledge, none of them have so far considered this unique property of course allocations: students want to enroll in courses together with their friends. 

In course allocation problems, each student must enroll in a set of courses. Students state their preferences over courses, and there exists a limit to the number of available course seats. In our setting, students also state the friends they would like to enroll in the course with. 
Each student is represented by a separate agent (e.g., the student's smartphone); this information is internal, i.e., only available to the agent. 

We formulated the problem as an asymmetric DCOP and developed a novel dedicated DCOP algorithm, \dsarc.

We have evaluated our algorithm both on simulated data and on data from a user study we conducted. Our \dsarc algorithm commonly finds valid solutions with high social welfare. The algorithm also maintains fairness in order that stems from its mode of operation; the fairness is reflected by the utilities of different agents (first, middle, and last). Moreover, the algorithm maintains equality, as expressed by the Gini coefficient.


In addition, we observed a correlation between the friendships' strength and our algorithm's performance. This provides a significant advantage to students who prioritize course choices based on friendships over the contents of the course itself. Furthermore, the \dsarc algorithm performs better in the user study than on the simulated data. This finding is very encouraging since we believe that the user study better reflects real-world problems than the simulated data.

Although the algorithm is not theoretically optimal, it fulfills the required course limitation even amid very constrained scenarios while providing a solution with high utility for the agents. 
Additionally, \dsarc is lightweight. Thus, it can be used to solve large course allocation problems in a timely manner. Although designed specifically for the problem of course allocation with friendships, \dsarc can be applied to other DCOPs and ADCOPs with resource capacity constraints.

The (A)DCOP is an inherently decentralized model, meaning that the student preferences are not explicitly stated to any single coordinator. This feature adds a layer of privacy that may incentivize students to state their truthful preferences. Nevertheless, some private information may leak throughout the solving process since agents send their values to their peers. The choice of a value incorporates implicit information on course preferences, friendship ratings, and course capacity constraints. Consequently, agents receiving such values may attempt to infer information about the private preferences and ratings. To tackle such privacy concerns, various privacy-preserving DCOP algorithms were developed in recent years~\citep{leaute2013protecting,grinshpoun2016psyncbb,tassa2017privacy,grinshpoun2019privacy}, including even immunity against colluding agents~\citep{tassa2021pcsyncbb,kogan2023privacy}. However, all these algorithms are computationally heavy as none uses local search. Thus, we consider the development of a privacy-preserving local search algorithm as an interesting and important direction for future research.

Another interesting future work direction is to devise a two-stage method in which $\dsarc$ is applied as the first stage. After the DCOP finds a solution with high social welfare, that solution can be the input of a second-stage mechanism that enforces some desired property, e.g., incentive compatibility. For instance, an adaptation of the Yankee swap mechanism~\citep{viswanathan2023yankee} to the course allocation domain can achieve strategyproofness. 

Finally, while the proposed solution is based on a course allocation draft mechanism in which students submit strict preferences, our model and algorithm can be easily modified to accept bids instead of strict preferences.



\section*{Acknowledgments}
The authors were supported by the Ariel Cyber Innovation Center in conjunction with the Israel National Cyber Directorate in the Prime Minister's Office.


\begin{thebibliography}{60}
\expandafter\ifx\csname natexlab\endcsname\relax\def\natexlab#1{#1}\fi
\providecommand{\url}[1]{\texttt{#1}}
\providecommand{\href}[2]{#2}
\providecommand{\path}[1]{#1}
\providecommand{\DOIprefix}{doi:}
\providecommand{\ArXivprefix}{arXiv:}
\providecommand{\URLprefix}{URL: }
\providecommand{\Pubmedprefix}{pmid:}
\providecommand{\doi}[1]{\href{http://dx.doi.org/#1}{\path{#1}}}
\providecommand{\Pubmed}[1]{\href{pmid:#1}{\path{#1}}}
\providecommand{\bibinfo}[2]{#2}
\ifx\xfnm\relax \def\xfnm[#1]{\unskip,\space#1}\fi
\bibitem[{Abdulkadiro{\u{g}}lu \& S{\"o}nmez(2003)}]{abdulkadirouglu2003school}
\bibinfo{author}{Abdulkadiro{\u{g}}lu, A.}, \& \bibinfo{author}{S{\"o}nmez, T.}
  (\bibinfo{year}{2003}).
\newblock \bibinfo{title}{School choice: A mechanism design approach}.
\newblock {\it \bibinfo{journal}{American economic review}\/},  {\it
  \bibinfo{volume}{93}\/}, \bibinfo{pages}{729--747}.
\bibitem[{Amanatidis et~al.(2017)Amanatidis, Birmpas, Christodoulou \&
  Markakis}]{amanatidis2017truthful}
\bibinfo{author}{Amanatidis, G.}, \bibinfo{author}{Birmpas, G.},
  \bibinfo{author}{Christodoulou, G.}, \& \bibinfo{author}{Markakis, E.}
  (\bibinfo{year}{2017}).
\newblock \bibinfo{title}{Truthful allocation mechanisms without payments:
  Characterization and implications on fairness}.
\newblock In {\it \bibinfo{booktitle}{Proceedings of the 2017 ACM Conference on
  Economics and Computation}\/} (pp. \bibinfo{pages}{545--562}).
\bibitem[{Amanatidis et~al.(2021)Amanatidis, Birmpas, Fusco, Lazos, Leonardi \&
  Reiffenh{\"a}user}]{amanatidis2021allocating}
\bibinfo{author}{Amanatidis, G.}, \bibinfo{author}{Birmpas, G.},
  \bibinfo{author}{Fusco, F.}, \bibinfo{author}{Lazos, P.},
  \bibinfo{author}{Leonardi, S.}, \& \bibinfo{author}{Reiffenh{\"a}user, R.}
  (\bibinfo{year}{2021}).
\newblock \bibinfo{title}{Allocating indivisible goods to strategic agents:
  Pure {Nash} equilibria and fairness}.
\newblock In {\it \bibinfo{booktitle}{International Conference on Web and
  Internet Economics}\/} (pp. \bibinfo{pages}{149--166}).
\newblock \bibinfo{organization}{Springer}.
\bibitem[{Atef~Yekta \& Day(2020)}]{yekta2020}
\bibinfo{author}{Atef~Yekta, H.}, \& \bibinfo{author}{Day, R.}
  (\bibinfo{year}{2020}).
\newblock \bibinfo{title}{Optimization-based mechanisms for the course
  allocation problem}.
\newblock {\it \bibinfo{journal}{INFORMS Journal on Computing}\/},  {\it
  \bibinfo{volume}{32}\/}, \bibinfo{pages}{641--660}.
\bibitem[{Azevedo \& Budish(2019)}]{azevedo2019strategy}
\bibinfo{author}{Azevedo, E.~M.}, \& \bibinfo{author}{Budish, E.}
  (\bibinfo{year}{2019}).
\newblock \bibinfo{title}{Strategy-proofness in the large}.
\newblock {\it \bibinfo{journal}{The Review of Economic Studies}\/},  {\it
  \bibinfo{volume}{86}\/}, \bibinfo{pages}{81--116}.
\bibitem[{Bichler et~al.(2021)Bichler, Hammerl, Morrill \&
  Waldherr}]{bichler2021assign}
\bibinfo{author}{Bichler, M.}, \bibinfo{author}{Hammerl, A.},
  \bibinfo{author}{Morrill, T.}, \& \bibinfo{author}{Waldherr, S.}
  (\bibinfo{year}{2021}).
\newblock \bibinfo{title}{How to assign scarce resources without money:
  Designing information systems that are efficient, truthful, and (pretty)
  fair}.
\newblock {\it \bibinfo{journal}{Information Systems Research}\/},  {\it
  \bibinfo{volume}{32}\/}, \bibinfo{pages}{335--355}.
\bibitem[{Bir{\'o}(2008)}]{biro2008student}
\bibinfo{author}{Bir{\'o}, P.} (\bibinfo{year}{2008}).
\newblock \bibinfo{title}{Student admissions in hungary as gale and shapley
  envisaged}.
\newblock {\it \bibinfo{journal}{University of Glasgow Technical Report
  TR-2008-291}\/}, .
\bibitem[{Budish(2011)}]{budish2011combinatorial}
\bibinfo{author}{Budish, E.} (\bibinfo{year}{2011}).
\newblock \bibinfo{title}{The combinatorial assignment problem: Approximate
  competitive equilibrium from equal incomes}.
\newblock {\it \bibinfo{journal}{Journal of Political Economy}\/},  {\it
  \bibinfo{volume}{119}\/}, \bibinfo{pages}{1061--1103}.
\bibitem[{Budish et~al.(2017)Budish, Cachon, Kessler \&
  Othman}]{budish2017course}
\bibinfo{author}{Budish, E.}, \bibinfo{author}{Cachon, G.~P.},
  \bibinfo{author}{Kessler, J.~B.}, \& \bibinfo{author}{Othman, A.}
  (\bibinfo{year}{2017}).
\newblock \bibinfo{title}{Course match: A large-scale implementation of
  approximate competitive equilibrium from equal incomes for combinatorial
  allocation}.
\newblock {\it \bibinfo{journal}{Operations Research}\/},  {\it
  \bibinfo{volume}{65}\/}, \bibinfo{pages}{314--336}.
\bibitem[{Budish \& Cantillon(2012)}]{budish2012multi}
\bibinfo{author}{Budish, E.}, \& \bibinfo{author}{Cantillon, E.}
  (\bibinfo{year}{2012}).
\newblock \bibinfo{title}{The multi-unit assignment problem: Theory and
  evidence from course allocation at {Harvard}}.
\newblock {\it \bibinfo{journal}{American Economic Review}\/},  {\it
  \bibinfo{volume}{102}\/}, \bibinfo{pages}{2237--71}.
\bibitem[{Budish \& Kessler(2022)}]{budish2022can}
\bibinfo{author}{Budish, E.}, \& \bibinfo{author}{Kessler, J.~B.}
  (\bibinfo{year}{2022}).
\newblock \bibinfo{title}{Can market participants report their preferences
  accurately (enough)?}
\newblock {\it \bibinfo{journal}{Management Science}\/},  {\it
  \bibinfo{volume}{68}\/}, \bibinfo{pages}{1107--1130}.
\bibitem[{Caragiannis et~al.(2019)Caragiannis, Kurokawa, Moulin, Procaccia,
  Shah \& Wang}]{caragiannis2019unreasonable}
\bibinfo{author}{Caragiannis, I.}, \bibinfo{author}{Kurokawa, D.},
  \bibinfo{author}{Moulin, H.}, \bibinfo{author}{Procaccia, A.~D.},
  \bibinfo{author}{Shah, N.}, \& \bibinfo{author}{Wang, J.}
  (\bibinfo{year}{2019}).
\newblock \bibinfo{title}{The unreasonable fairness of maximum nash welfare}.
\newblock {\it \bibinfo{journal}{ACM Transactions on Economics and Computation
  (TEAC)}\/},  {\it \bibinfo{volume}{7}\/}, \bibinfo{pages}{1--32}.
\bibitem[{Carley \& Krackhardt(1996)}]{carley1996cognitive}
\bibinfo{author}{Carley, K.~M.}, \& \bibinfo{author}{Krackhardt, D.}
  (\bibinfo{year}{1996}).
\newblock \bibinfo{title}{Cognitive inconsistencies and non-symmetric
  friendship}.
\newblock {\it \bibinfo{journal}{Social networks}\/},  {\it
  \bibinfo{volume}{18}\/}, \bibinfo{pages}{1--27}.
\bibitem[{Delacr{\'e}taz et~al.(2016)Delacr{\'e}taz, Kominers \&
  Teytelboym}]{delacretaz2016refugee}
\bibinfo{author}{Delacr{\'e}taz, D.}, \bibinfo{author}{Kominers, S.~D.}, \&
  \bibinfo{author}{Teytelboym, A.} (\bibinfo{year}{2016}).
\newblock \bibinfo{title}{Refugee resettlement}.
\newblock {\it \bibinfo{journal}{University of Oxford Department of Economics
  Working Paper}\/}, .
\bibitem[{Diebold et~al.(2014)Diebold, Aziz, Bichler, Matthes \&
  Schneider}]{diebold2014course}
\bibinfo{author}{Diebold, F.}, \bibinfo{author}{Aziz, H.},
  \bibinfo{author}{Bichler, M.}, \bibinfo{author}{Matthes, F.}, \&
  \bibinfo{author}{Schneider, A.} (\bibinfo{year}{2014}).
\newblock \bibinfo{title}{Course allocation via stable matching}.
\newblock {\it \bibinfo{journal}{Business \& Information Systems
  Engineering}\/},  {\it \bibinfo{volume}{6}\/}, \bibinfo{pages}{97--110}.
\bibitem[{Ehlers \& Klaus(2003)}]{ehlers2003coalitional}
\bibinfo{author}{Ehlers, L.}, \& \bibinfo{author}{Klaus, B.}
  (\bibinfo{year}{2003}).
\newblock \bibinfo{title}{Coalitional strategy-proof and resource-monotonic
  solutions for multiple assignment problems}.
\newblock {\it \bibinfo{journal}{Social Choice and Welfare}\/},  {\it
  \bibinfo{volume}{21}\/}, \bibinfo{pages}{265--280}.
\bibitem[{Elkind et~al.(2020)Elkind, Patel, Tsang \& Zick}]{elkindPTZ20}
\bibinfo{author}{Elkind, E.}, \bibinfo{author}{Patel, N.},
  \bibinfo{author}{Tsang, A.}, \& \bibinfo{author}{Zick, Y.}
  (\bibinfo{year}{2020}).
\newblock \bibinfo{title}{Keeping your friends close: Land allocation with
  friends}.
\newblock In {\it \bibinfo{booktitle}{IJCAI}\/} (pp.
  \bibinfo{pages}{318--324}).
\bibitem[{Farinelli et~al.(2008)Farinelli, Rogers, Petcu \&
  Jennings}]{farinelli2008decentralised}
\bibinfo{author}{Farinelli, A.}, \bibinfo{author}{Rogers, A.},
  \bibinfo{author}{Petcu, A.}, \& \bibinfo{author}{Jennings, N.~R.}
  (\bibinfo{year}{2008}).
\newblock \bibinfo{title}{Decentralised coordination of low-power embedded
  devices using the {Max-Sum} algorithm}.
\newblock In {\it \bibinfo{booktitle}{AAMAS}\/} (pp.
  \bibinfo{pages}{639--646}).
\bibitem[{Fioretto et~al.(2018)Fioretto, Pontelli \&
  Yeoh}]{fioretto2018distributed}
\bibinfo{author}{Fioretto, F.}, \bibinfo{author}{Pontelli, E.}, \&
  \bibinfo{author}{Yeoh, W.} (\bibinfo{year}{2018}).
\newblock \bibinfo{title}{Distributed constraint optimization problems and
  applications: A survey}.
\newblock {\it \bibinfo{journal}{Journal of Artificial Intelligence
  Research}\/},  {\it \bibinfo{volume}{61}\/}, \bibinfo{pages}{623--698}.
\bibitem[{Foley(1966)}]{foley1966resource}
\bibinfo{author}{Foley, D.~K.} (\bibinfo{year}{1966}).
\newblock {\it \bibinfo{title}{Resource allocation and the public sector}\/}.
\newblock \bibinfo{publisher}{Yale University}.
\bibitem[{Gale \& Shapley(1962)}]{gale1962college}
\bibinfo{author}{Gale, D.}, \& \bibinfo{author}{Shapley, L.~S.}
  (\bibinfo{year}{1962}).
\newblock \bibinfo{title}{College admissions and the stability of marriage}.
\newblock {\it \bibinfo{journal}{The American Mathematical Monthly}\/},  {\it
  \bibinfo{volume}{69}\/}, \bibinfo{pages}{9--15}.
\bibitem[{Gershman et~al.(2008)Gershman, Grubshtein, Meisels, Rokach \&
  Zivan}]{gershman2008scheduling}
\bibinfo{author}{Gershman, A.}, \bibinfo{author}{Grubshtein, A.},
  \bibinfo{author}{Meisels, A.}, \bibinfo{author}{Rokach, L.}, \&
  \bibinfo{author}{Zivan, R.} (\bibinfo{year}{2008}).
\newblock \bibinfo{title}{Scheduling meetings by agents}.
\newblock In {\it \bibinfo{booktitle}{PATAT}\/}.
\bibitem[{Gini(1936)}]{gini1936measure}
\bibinfo{author}{Gini, C.} (\bibinfo{year}{1936}).
\newblock \bibinfo{title}{On the measure of concentration with special
  reference to income and statistics}.
\newblock {\it \bibinfo{journal}{Colorado College Publication, General
  Series}\/},  {\it \bibinfo{volume}{208}\/}, \bibinfo{pages}{73--79}.
\bibitem[{Gourv{\`e}s et~al.(2014)Gourv{\`e}s, Monnot \&
  Tlilane}]{gourves2014near}
\bibinfo{author}{Gourv{\`e}s, L.}, \bibinfo{author}{Monnot, J.}, \&
  \bibinfo{author}{Tlilane, L.} (\bibinfo{year}{2014}).
\newblock \bibinfo{title}{Near fairness in matroids.}
\newblock In {\it \bibinfo{booktitle}{ECAI}\/} (pp. \bibinfo{pages}{393--398}).
\bibitem[{Grinshpoun et~al.(2013)Grinshpoun, Grubshtein, Zivan, Netzer \&
  Meisels}]{grinshpoun2013asymmetric}
\bibinfo{author}{Grinshpoun, T.}, \bibinfo{author}{Grubshtein, A.},
  \bibinfo{author}{Zivan, R.}, \bibinfo{author}{Netzer, A.}, \&
  \bibinfo{author}{Meisels, A.} (\bibinfo{year}{2013}).
\newblock \bibinfo{title}{Asymmetric distributed constraint optimization
  problems}.
\newblock {\it \bibinfo{journal}{Journal of Artificial Intelligence
  Research}\/},  {\it \bibinfo{volume}{47}\/}, \bibinfo{pages}{613--647}.
\bibitem[{Grinshpoun \& Tassa(2016)}]{grinshpoun2016psyncbb}
\bibinfo{author}{Grinshpoun, T.}, \& \bibinfo{author}{Tassa, T.}
  (\bibinfo{year}{2016}).
\newblock \bibinfo{title}{{P-SyncBB}: {A} privacy preserving branch and bound
  {DCOP} algorithm}.
\newblock {\it \bibinfo{journal}{Journal of Artificial Intelligence
  Research}\/},  {\it \bibinfo{volume}{57}\/}, \bibinfo{pages}{621--660}.
\bibitem[{Grinshpoun et~al.(2019)Grinshpoun, Tassa, Levit \&
  Zivan}]{grinshpoun2019privacy}
\bibinfo{author}{Grinshpoun, T.}, \bibinfo{author}{Tassa, T.},
  \bibinfo{author}{Levit, V.}, \& \bibinfo{author}{Zivan, R.}
  (\bibinfo{year}{2019}).
\newblock \bibinfo{title}{Privacy preserving region optimal algorithms for
  symmetric and asymmetric {DCOPs}}.
\newblock {\it \bibinfo{journal}{Artificial Intelligence}\/},  {\it
  \bibinfo{volume}{266}\/}, \bibinfo{pages}{27--50}.
\bibitem[{Hackett et~al.(2018)Hackett, Johnston \&
  Bilen}]{hackett2018spacecraft}
\bibinfo{author}{Hackett, T.~M.}, \bibinfo{author}{Johnston, M.}, \&
  \bibinfo{author}{Bilen, S.~G.} (\bibinfo{year}{2018}).
\newblock \bibinfo{title}{Spacecraft block scheduling for nasa’s deep space
  network}.
\newblock In {\it \bibinfo{booktitle}{2018 SpaceOps Conference}\/} (p.
  \bibinfo{pages}{2578}).
\bibitem[{Hatfield(2009)}]{hatfield2009strategy}
\bibinfo{author}{Hatfield, J.~W.} (\bibinfo{year}{2009}).
\newblock \bibinfo{title}{Strategy-proof, efficient, and nonbossy quota
  allocations}.
\newblock {\it \bibinfo{journal}{Social Choice and Welfare}\/},  {\it
  \bibinfo{volume}{33}\/}, \bibinfo{pages}{505--515}.
\bibitem[{Johnston \& Wyatt(2017)}]{johnston2017ai}
\bibinfo{author}{Johnston, M.~D.}, \& \bibinfo{author}{Wyatt, E.~J.}
  (\bibinfo{year}{2017}).
\newblock \bibinfo{title}{{AI} and autonomy initiatives for {NASA}’s deep
  space network ({DSN})}, .
\bibitem[{Juthamanee et~al.(2021)Juthamanee, Piromsopa \&
  Chongstitvatana}]{juthamanee2021token}
\bibinfo{author}{Juthamanee, C.}, \bibinfo{author}{Piromsopa, K.}, \&
  \bibinfo{author}{Chongstitvatana, P.} (\bibinfo{year}{2021}).
\newblock \bibinfo{title}{Token allocation for course bidding with machine
  learning method}.
\newblock In {\it \bibinfo{booktitle}{2021 18th International Conference on
  Electrical Engineering/Electronics, Computer, Telecommunications and
  Information Technology (ECTI-CON)}\/} (pp. \bibinfo{pages}{1168--1171}).
\newblock \bibinfo{organization}{IEEE}.
\bibitem[{Khakhiashvili et~al.(2021)Khakhiashvili, Grinshpoun \&
  Dery}]{khakhiashvili2021course}
\bibinfo{author}{Khakhiashvili, I.}, \bibinfo{author}{Grinshpoun, T.}, \&
  \bibinfo{author}{Dery, L.} (\bibinfo{year}{2021}).
\newblock \bibinfo{title}{Course allocation with friendships as an asymmetric
  distributed constraint optimization problem}.
\newblock In {\it \bibinfo{booktitle}{IEEE/WIC/ACM International Conference on
  Web Intelligence and Intelligent Agent Technology}\/} (pp.
  \bibinfo{pages}{688--693}).
\bibitem[{Klaus \& Miyagawa(2002)}]{klaus2002strategy}
\bibinfo{author}{Klaus, B.}, \& \bibinfo{author}{Miyagawa, E.}
  (\bibinfo{year}{2002}).
\newblock \bibinfo{title}{Strategy-proofness, solidarity, and consistency for
  multiple assignment problems}.
\newblock {\it \bibinfo{journal}{International Journal of Game Theory}\/},
  {\it \bibinfo{volume}{30}\/}, \bibinfo{pages}{421--435}.
\bibitem[{Kogan et~al.(2023)Kogan, Tassa \& Grinshpoun}]{kogan2023privacy}
\bibinfo{author}{Kogan, P.}, \bibinfo{author}{Tassa, T.}, \&
  \bibinfo{author}{Grinshpoun, T.} (\bibinfo{year}{2023}).
\newblock \bibinfo{title}{Privacy preserving solution of {DCOPs} by mediation}.
\newblock {\it \bibinfo{journal}{Artificial Intelligence}\/},  (p.
  \bibinfo{pages}{103916}).
\bibitem[{Kominers et~al.(2010)Kominers, Ruberry \&
  Ullman}]{kominers2010course}
\bibinfo{author}{Kominers, S.~D.}, \bibinfo{author}{Ruberry, M.}, \&
  \bibinfo{author}{Ullman, J.} (\bibinfo{year}{2010}).
\newblock \bibinfo{title}{Course allocation by proxy auction}.
\newblock In {\it \bibinfo{booktitle}{International Workshop on Internet and
  Network Economics}\/} (pp. \bibinfo{pages}{551--558}).
\newblock \bibinfo{organization}{Springer}.
\bibitem[{Krishna \& {\"U}nver(2008)}]{krishna2008research}
\bibinfo{author}{Krishna, A.}, \& \bibinfo{author}{{\"U}nver, M.~U.}
  (\bibinfo{year}{2008}).
\newblock \bibinfo{title}{Research note—improving the efficiency of course
  bidding at business schools: Field and laboratory studies}.
\newblock {\it \bibinfo{journal}{Marketing Science}\/},  {\it
  \bibinfo{volume}{27}\/}, \bibinfo{pages}{262--282}.
\bibitem[{Lancieri(2017)}]{lancieri2017asymmetry}
\bibinfo{author}{Lancieri, L.} (\bibinfo{year}{2017}).
\newblock \bibinfo{title}{Asymmetry in the perception of friendship in students
  groups.}
\newblock {\it \bibinfo{journal}{International Association for Development of
  the Information Society}\/}, .
\bibitem[{L{\'e}aut{\'e} \& Faltings(2011)}]{leaute2011distributed}
\bibinfo{author}{L{\'e}aut{\'e}, T.}, \& \bibinfo{author}{Faltings, B.}
  (\bibinfo{year}{2011}).
\newblock \bibinfo{title}{Distributed constraint optimization under stochastic
  uncertainty}.
\newblock In {\it \bibinfo{booktitle}{AAAI}\/} (pp. \bibinfo{pages}{68--73}).
\bibitem[{L{\'e}aut{\'e} \& Faltings(2013)}]{leaute2013protecting}
\bibinfo{author}{L{\'e}aut{\'e}, T.}, \& \bibinfo{author}{Faltings, B.}
  (\bibinfo{year}{2013}).
\newblock \bibinfo{title}{Protecting privacy through distributed computation in
  multi-agent decision making}.
\newblock {\it \bibinfo{journal}{Journal of Artificial Intelligence
  Research}\/},  {\it \bibinfo{volume}{47}\/}, \bibinfo{pages}{649--695}.
\bibitem[{Lezama et~al.(2019)Lezama, Palominos, Rodr{\'\i}guez-Gonz{\'a}lez,
  Farinelli \& de~Cote}]{lezama2019agent}
\bibinfo{author}{Lezama, F.}, \bibinfo{author}{Palominos, J.},
  \bibinfo{author}{Rodr{\'\i}guez-Gonz{\'a}lez, A.~Y.},
  \bibinfo{author}{Farinelli, A.}, \& \bibinfo{author}{de~Cote, E.~M.}
  (\bibinfo{year}{2019}).
\newblock \bibinfo{title}{Agent-based microgrid scheduling: An {ICT}
  perspective}.
\newblock {\it \bibinfo{journal}{Mobile Networks and Applications}\/},  {\it
  \bibinfo{volume}{24}\/}, \bibinfo{pages}{1682--1698}.
\bibitem[{Lipton et~al.(2004)Lipton, Markakis, Mossel \&
  Saberi}]{Lipton2004EF1}
\bibinfo{author}{Lipton, R.~J.}, \bibinfo{author}{Markakis, E.},
  \bibinfo{author}{Mossel, E.}, \& \bibinfo{author}{Saberi, A.}
  (\bibinfo{year}{2004}).
\newblock \bibinfo{title}{On approximately fair allocations of indivisible
  goods}.
\newblock In {\it \bibinfo{booktitle}{Proceedings of the 5th International
  Conference on Economics and Computation (EC)}\/} (p.
  \bibinfo{pages}{125–131}).
\bibitem[{Lutati et~al.(2014)Lutati, Gontmakher, Lando, Netzer, Meisels \&
  Grubshtein}]{lutati2014agentzero}
\bibinfo{author}{Lutati, B.}, \bibinfo{author}{Gontmakher, I.},
  \bibinfo{author}{Lando, M.}, \bibinfo{author}{Netzer, A.},
  \bibinfo{author}{Meisels, A.}, \& \bibinfo{author}{Grubshtein, A.}
  (\bibinfo{year}{2014}).
\newblock \bibinfo{title}{Agentzero: A framework for simulating and evaluating
  multi-agent algorithms}.
\newblock In {\it \bibinfo{booktitle}{Agent-Oriented Software Engineering}\/}
  (pp. \bibinfo{pages}{309--327}).
\newblock \bibinfo{organization}{Springer}.
\bibitem[{Maheswaran et~al.(2004)Maheswaran, Tambe, Bowring, Pearce \&
  Varakantham}]{maheswaran2004taking}
\bibinfo{author}{Maheswaran, R.~T.}, \bibinfo{author}{Tambe, M.},
  \bibinfo{author}{Bowring, E.}, \bibinfo{author}{Pearce, J.~P.}, \&
  \bibinfo{author}{Varakantham, P.} (\bibinfo{year}{2004}).
\newblock \bibinfo{title}{Taking {DCOP} to the real world: Efficient complete
  solutions for distributed multi-event scheduling}.
\newblock In {\it \bibinfo{booktitle}{Proceedings of the 3rd International
  Conference on Autonomous Agents and Multiagent Systems (AAMAS)}\/} (pp.
  \bibinfo{pages}{310--317}).
\newblock \bibinfo{organization}{ACM}.
\bibitem[{Mattei \& Walsh(2013)}]{MaWa13a}
\bibinfo{author}{Mattei, N.}, \& \bibinfo{author}{Walsh, T.}
  (\bibinfo{year}{2013}).
\newblock \bibinfo{title}{Preflib: A library of preference data
  \textsc{http://preflib.org}}.
\newblock In {\it \bibinfo{booktitle}{ADT}\/} (pp. \bibinfo{pages}{259--270}).
\bibitem[{Modi et~al.(2005)Modi, Shen, Tambe \& Yokoo}]{modi2005adopt}
\bibinfo{author}{Modi, P.~J.}, \bibinfo{author}{Shen, W.-M.},
  \bibinfo{author}{Tambe, M.}, \& \bibinfo{author}{Yokoo, M.}
  (\bibinfo{year}{2005}).
\newblock \bibinfo{title}{Adopt: Asynchronous distributed constraint
  optimization with quality guarantees}.
\newblock {\it \bibinfo{journal}{Artificial Intelligence}\/},  {\it
  \bibinfo{volume}{161}\/}, \bibinfo{pages}{149--180}.
\bibitem[{Nogareda \& Camacho(2016)}]{nogareda2016optimizing}
\bibinfo{author}{Nogareda, A.-M.}, \& \bibinfo{author}{Camacho, D.}
  (\bibinfo{year}{2016}).
\newblock \bibinfo{title}{Optimizing satisfaction in a multi-courses allocation
  problem}.
\newblock In {\it \bibinfo{booktitle}{Intelligent Distributed Computing IX}\/}
  (pp. \bibinfo{pages}{247--256}).
\newblock \bibinfo{publisher}{Springer}.
\bibitem[{P{\'a}pai(2001)}]{papai2001strategyproof}
\bibinfo{author}{P{\'a}pai, S.} (\bibinfo{year}{2001}).
\newblock \bibinfo{title}{Strategyproof and nonbossy multiple assignments}.
\newblock {\it \bibinfo{journal}{Journal of Public Economic Theory}\/},  {\it
  \bibinfo{volume}{3}\/}, \bibinfo{pages}{257--271}.
\bibitem[{Roth(2002)}]{roth2002economist}
\bibinfo{author}{Roth, A.~E.} (\bibinfo{year}{2002}).
\newblock \bibinfo{title}{The economist as engineer: Game theory,
  experimentation, and computation as tools for design economics}.
\newblock {\it \bibinfo{journal}{Econometrica}\/},  {\it
  \bibinfo{volume}{70}\/}, \bibinfo{pages}{1341--1378}.
\bibitem[{Smith \& Peterson(2007)}]{smith2007psst}
\bibinfo{author}{Smith, R.~A.}, \& \bibinfo{author}{Peterson, B.~L.}
  (\bibinfo{year}{2007}).
\newblock \bibinfo{title}{“psst… what do you think?” the relationship
  between advice prestige, type of advice, and academic performance}.
\newblock {\it \bibinfo{journal}{Communication Education}\/},  {\it
  \bibinfo{volume}{56}\/}, \bibinfo{pages}{278--291}.
\bibitem[{S{\"o}nmez \& {\"U}nver(2010)}]{sonmez2010course}
\bibinfo{author}{S{\"o}nmez, T.}, \& \bibinfo{author}{{\"U}nver, M.~U.}
  (\bibinfo{year}{2010}).
\newblock \bibinfo{title}{Course bidding at business schools}.
\newblock {\it \bibinfo{journal}{International Economic Review}\/},  {\it
  \bibinfo{volume}{51}\/}, \bibinfo{pages}{99--123}.
\bibitem[{Sparrowe et~al.(2001)Sparrowe, Liden, Wayne \&
  Kraimer}]{sparrowe2001social}
\bibinfo{author}{Sparrowe, R.~T.}, \bibinfo{author}{Liden, R.~C.},
  \bibinfo{author}{Wayne, S.~J.}, \& \bibinfo{author}{Kraimer, M.~L.}
  (\bibinfo{year}{2001}).
\newblock \bibinfo{title}{Social networks and the performance of individuals
  and groups}.
\newblock {\it \bibinfo{journal}{Academy of management journal}\/},  {\it
  \bibinfo{volume}{44}\/}, \bibinfo{pages}{316--325}.
\bibitem[{Tassa et~al.(2021)Tassa, Grinshpoun \& Yanai}]{tassa2021pcsyncbb}
\bibinfo{author}{Tassa, T.}, \bibinfo{author}{Grinshpoun, T.}, \&
  \bibinfo{author}{Yanai, A.} (\bibinfo{year}{2021}).
\newblock \bibinfo{title}{{PC-SyncBB}: A privacy preserving collusion secure
  {DCOP} algorithm}.
\newblock {\it \bibinfo{journal}{Artificial Intelligence}\/},  {\it
  \bibinfo{volume}{297}\/}, \bibinfo{pages}{103501}.
\bibitem[{Tassa et~al.(2017)Tassa, Grinshpoun \& Zivan}]{tassa2017privacy}
\bibinfo{author}{Tassa, T.}, \bibinfo{author}{Grinshpoun, T.}, \&
  \bibinfo{author}{Zivan, R.} (\bibinfo{year}{2017}).
\newblock \bibinfo{title}{Privacy preserving implementation of the {Max-Sum}
  algorithm and its variants}.
\newblock {\it \bibinfo{journal}{Journal of Artificial Intelligence
  Research}\/},  {\it \bibinfo{volume}{59}\/}, \bibinfo{pages}{311--349}.
\bibitem[{Viswanathan \& Zick(2023)}]{viswanathan2023yankee}
\bibinfo{author}{Viswanathan, V.}, \& \bibinfo{author}{Zick, Y.}
  (\bibinfo{year}{2023}).
\newblock \bibinfo{title}{Yankee swap: A fast and simple fair allocation
  mechanism for matroid rank valuations}.
\newblock In {\it \bibinfo{booktitle}{Proceedings of the 2023 International
  Conference on Autonomous Agents and Multiagent Systems}\/} (pp.
  \bibinfo{pages}{179--187}).
\bibitem[{Wang \& Zhang(2022)}]{wang2022task}
\bibinfo{author}{Wang, Z.}, \& \bibinfo{author}{Zhang, J.}
  (\bibinfo{year}{2022}).
\newblock \bibinfo{title}{A task allocation algorithm for a swarm of unmanned
  aerial vehicles based on bionic wolf pack method}.
\newblock {\it \bibinfo{journal}{Knowledge-Based Systems}\/},  {\it
  \bibinfo{volume}{250}\/}, \bibinfo{pages}{109072}.
\bibitem[{Witkow \& Fuligni(2010)}]{witkow2010school}
\bibinfo{author}{Witkow, M.~R.}, \& \bibinfo{author}{Fuligni, A.~J.}
  (\bibinfo{year}{2010}).
\newblock \bibinfo{title}{In-school versus out-of-school friendships and
  academic achievement among an ethnically diverse sample of adolescents}.
\newblock {\it \bibinfo{journal}{Journal of research on adolescence}\/},  {\it
  \bibinfo{volume}{20}\/}, \bibinfo{pages}{631--650}.
\bibitem[{Zhang et~al.(2005)Zhang, Wang, Xing \&
  Wittenburg}]{zhang2005distributed}
\bibinfo{author}{Zhang, W.}, \bibinfo{author}{Wang, G.}, \bibinfo{author}{Xing,
  Z.}, \& \bibinfo{author}{Wittenburg, L.} (\bibinfo{year}{2005}).
\newblock \bibinfo{title}{Distributed stochastic search and distributed
  breakout: properties, comparison and applications to constraint optimization
  problems in sensor networks}.
\newblock {\it \bibinfo{journal}{Artificial Intelligence}\/},  {\it
  \bibinfo{volume}{161}\/}, \bibinfo{pages}{55--87}.
\bibitem[{Zhao et~al.(2023)Zhao, Dong, Wang \& Pan}]{zhao2023ppo}
\bibinfo{author}{Zhao, B.}, \bibinfo{author}{Dong, H.}, \bibinfo{author}{Wang,
  Y.}, \& \bibinfo{author}{Pan, T.} (\bibinfo{year}{2023}).
\newblock \bibinfo{title}{Ppo-ta: Adaptive task allocation via proximal policy
  optimization for spatio-temporal crowdsourcing}.
\newblock {\it \bibinfo{journal}{Knowledge-Based Systems}\/},  (p.
  \bibinfo{pages}{110330}).
\bibitem[{Zilberstein(1996)}]{zilberstein1996using}
\bibinfo{author}{Zilberstein, S.} (\bibinfo{year}{1996}).
\newblock \bibinfo{title}{Using anytime algorithms in intelligent systems}.
\newblock {\it \bibinfo{journal}{AI Magazine}\/},  {\it
  \bibinfo{volume}{17}\/}, \bibinfo{pages}{73--73}.
\bibitem[{Zivan et~al.(2014)Zivan, Okamoto \& Peled}]{zivan2014explorative}
\bibinfo{author}{Zivan, R.}, \bibinfo{author}{Okamoto, S.}, \&
  \bibinfo{author}{Peled, H.} (\bibinfo{year}{2014}).
\newblock \bibinfo{title}{Explorative anytime local search for distributed
  constraint optimization}.
\newblock {\it \bibinfo{journal}{Artificial Intelligence}\/},  {\it
  \bibinfo{volume}{212}\/}, \bibinfo{pages}{1--26}.

\end{thebibliography}

\end{document}